\documentclass[letterpaper, 10 pt, conference]{IEEEtran} 
\IEEEoverridecommandlockouts                          
\usepackage{times}

\usepackage[numbers]{natbib}
\usepackage{multicol}
\usepackage[bookmarks=true]{hyperref}
\usepackage{graphicx}
\usepackage{caption}
\usepackage{multirow}
\usepackage{amsmath} % assumes amsmath package installed
\usepackage{amssymb}  % assumes amsmath package installed
\usepackage{soul}
\usepackage{array}
\usepackage[table,xcdraw,dvipsnames]{xcolor}
\usepackage{wrapfig}
\usepackage{subcaption}
\let\labelindent\relax
\usepackage{enumitem}
\usepackage{algorithm}
\usepackage[noend]{algpseudocode}
\usepackage{bbding}
\usepackage{pifont}
\usepackage{wasysym}
\usepackage{algorithm, algpseudocode}
\usepackage{boldline}

\usepackage{capt-of,etoolbox}

\newenvironment{flushitemize}{
  \begin{itemize}[leftmargin=*, labelsep=0.5em]
    \setlength{\itemindent}{0pt}
    \setlength{\labelwidth}{0pt}
    \setlength{\labelsep}{0.5em}
}{\end{itemize}}

\begin{document}

% paper title
\title{Local Solution Repair for Multi-Agent Pickup and Delivery}

\author{Ethan Schneider$^{1}$, Jingkai Chen$^{2}$, Seth Hutchinson$^{3}$, Sonia Chernova$^{1}$% <-this % stops a space
% \thanks{\NEW{Manuscript received: }}
% \thanks{\NEW{This paper was recommended for publication by ...}}
\thanks{This work is sponsored by Symbotic.}% <-this % stops a space
\thanks{$^{1}$Institute of Robotics and Intelligent Machines, Georgia Institute of Technology, Atlanta, GA 30332, USA \texttt{eschneider32@gatech.edu}}%
\thanks{$^{2}$Symbotic Inc., Wilmington, MA 01887, USA}%
\thanks{$^{3}$Northeastern University, Boston, MA 02115, USA}%
}

\maketitle

\begin{abstract}
Multi-robot systems in automated warehouses must efficiently attend to a constant stream of pickup-and-delivery tasks while avoiding collisions. Existing Multi-Agent Pickup-and-Delivery (MAPD) solutions generate plans that are globally feasible but often locally inefficient. In large-scale, long-horizon deployments, these inefficiencies, such as robots backtracking from target goal locations and idling in congested areas, accumulate over time and degrade performance. This paper contributes Local Solution Repair (LSR), a goal replanning approach that detects suboptimal behaviors at run-time and locally improves plans to reduce conflict between agents. We demonstrate the utility of our approach by repairing solutions generated by state-of-the-art MAPD baselines, M2M and LNS-PBS. Our results in long-horizon warehouse simulations over three environments show upwards of 5.9\% improvement in task throughput over current state-of-the-art methods by reducing the number of suboptimal instances by $\approx$36k or by 14.6\% with minor additional computation time.
\end{abstract}

\IEEEpeerreviewmaketitle

\section{Introduction}
\label{sec:introduction}
\begin{figure}[t!]
    \centering
    \includegraphics[width=0.9\linewidth]{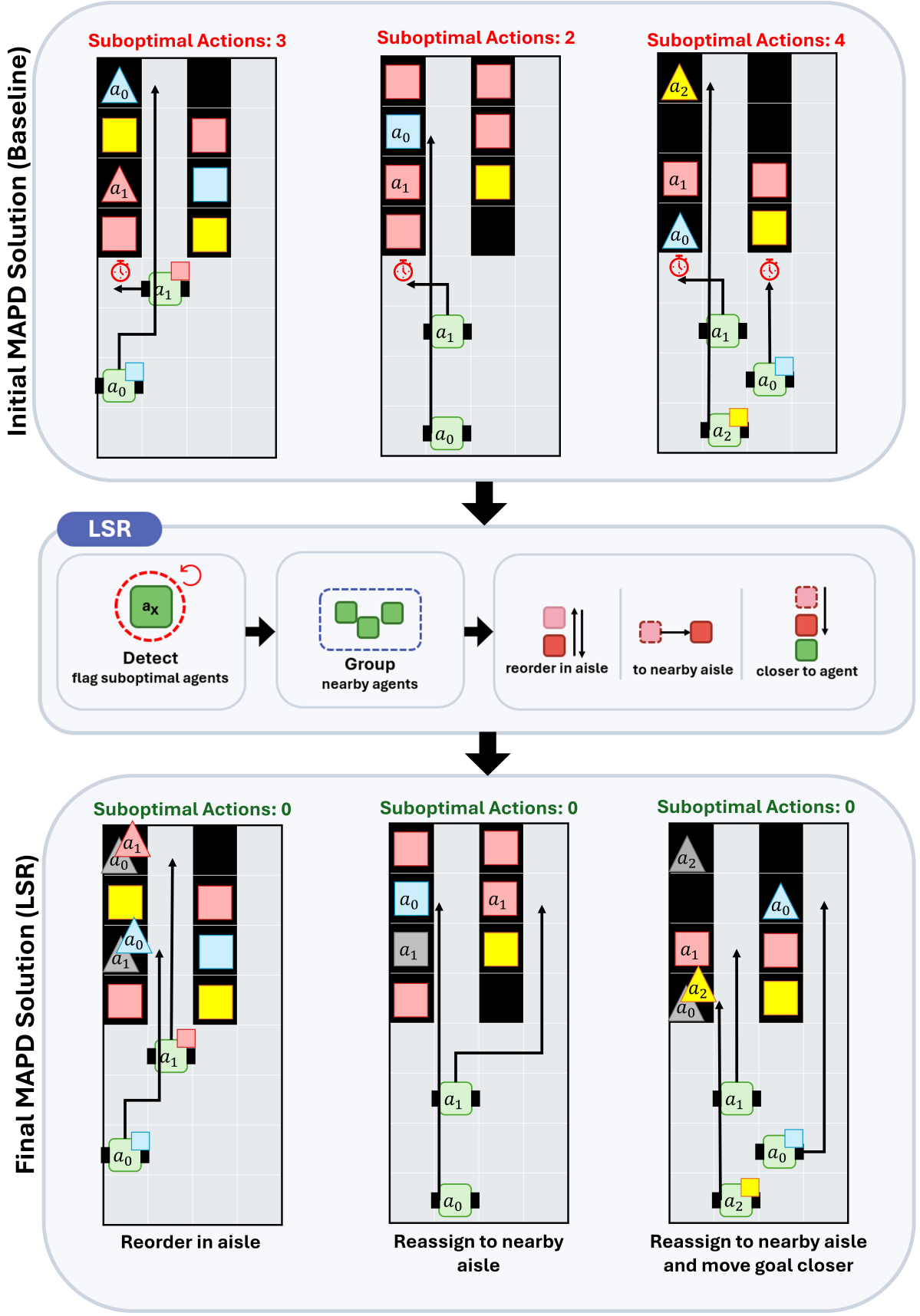}
    \caption{\small Examples of initial MAPD solutions in which robots pick up objects $\square$ and deliver them to destinations $\triangle$ within a warehouse scenario. The three initial MAPD solutions show examples of suboptimal actions (e.g., idling and backtracking) due to suboptimal and/or collision-unaware task allocations. For each of these scenarios, \textbf{LSR} detects the agents executing suboptimal actions, groups other agents contributing to these behaviors, then reassigns goals locations for the subset of agents. The final MAPD solutions show that LSR improves the overall solutions by: \textbf{(1)} rearranging goal order within aisles, \textbf{(2)} moving goal(s) to neighboring aisles, \textbf{(3)} moving goal(s) closer to an agent's current location, or \textbf{(4)} a combination of the previous ways.\vspace{-1em}}
    \label{fig:motivation}
    \vspace{-1em}
\end{figure}

In automated warehouse logistics, teams of robots are deployed to organize warehouse inventory, store incoming items, and retrieve stock for shipping. The warehouse often contains multiple interchangeable instances of a given item (e.g., many bags of chips or boxes of vitamins) distributed across different locations, requiring the system to determine which specific instance to retrieve. Coordinating the selection, pickup, and delivery of items in response to such requests is formalized as the Many-to-Many Multi-Agent Pickup and Delivery (M2M-MAPD) \cite{schneider2026} problem, a variant of the Multi-Agent Pickup-and-Delivery (MAPD) \cite{ma2017lifelong} problem.
% , a known NP-hard problem \cite{ma2017lifelong}. 
To solve a MAPD problem, algorithmic solutions must allocate agents to tasks and compute motion plans to maximize throughput $\eta$, defined as the number of tasks completed per minute.

Computing MAPD solutions is a known NP-hard problem \cite{ma2017lifelong}, so prior MAPD methods sacrifice optimality for scalability and computational efficiency by sequentially computing a collision-unaware task allocation and then the corresponding motion plan \cite{LNS-PBS, schneider2026, liu2019task, ma2017lifelong}. Prior work has found success with task allocation methods such as Large Neighborhood Search (LNS) \cite{pisinger2018large} and scalable suboptimal path finding algorithms such as PBS \cite{ma2019searching} and EECBS \cite{barer2014suboptimal}, which produce scalable good solutions in practice \cite{LNS-PBS, schneider2026}.

% , using suboptimal task allocation algorithms (e.g., Large Neighborhood Search \cite{pisinger2018large}), or utilizing suboptimal path finding algorithms (e.g., PBS \cite{ma2019searching}, EECBS \cite{barer2014suboptimal}). 

While these design decisions enable prior MAPD methods to scale more efficiently, they can result in suboptimal agent behavior during runtime. Figure~\ref{fig:motivation} illustrates two common suboptimal behaviors, \textbf{backtracking}, in which an agent moves away from its current goal to yield to another agent and must subsequently retrace its path, and \textbf{idling}, in which an agent remains stationary despite having an active goal, waiting for another agent to pass. Both behaviors are necessary to resolve local conflicts, but represent \textit{suboptimal actions} that reduce throughput. Our results show that over multiple 8-hour simulations with 40 robots, prior methods such as LNS-PBS \cite{LNS-PBS} and M2M \cite{schneider2026} accumulate over $250$k suboptimal actions (out of $1.15$M actions), equivalent to $13$ suboptimal actions per robot per minute, decreasing overall system efficiency.  

Our goal is to identify and correct these inter-robot conflicts, which arise from collision-unaware cost estimates and suboptimal tasking or routing, without incurring the overhead of replanning the entire system. Rather than re-solving the full MAPD instance, we reassign only a subset of agents to new goal locations. This local repair is made possible by the M2M-MAPD formulation, in which each task has multiple interchangeable pickup and dropoff locations: an agent can be redirected to a different empty storage location when dropping off an item, or to a different copy of the requested item when picking one up. Because these alternatives are equivalent from the task's perspective, reassigning an agent's goal keeps the task valid while giving the system the freedom to resolve the local congestion that drives suboptimal actions.

In this paper, we formalize the \textbf{Suboptimality Detection and Repair (SDR) problem}, in which suboptimal behaviors produced by an existing MAPD solution are identified and repaired to improve overall system performance. To solve the SDR problem, we introduce \textbf{Local Solution Repair (LSR)}, which identifies subsets of agents that exhibit and contribute to suboptimal behavior. LSR then performs a local goal reassignment to alleviate the suboptimal behavior by reassigning agents' goals to nearby equivalent locations. We introduce two LSR variants to compute new goal assignments: Estimated-Cost Repair ($\text{LSR}_{\text{ECR}}$) uses estimated path costs to candidate goal locations and assigns them to the group of agents using the Hungarian Algorithm \cite{kuhn1955hungarian}, and MAPF-Cost Repair ($\text{LSR}_{\text{MCR}}$) uses a Branch-and-Bound algorithm to reassign the group of agents to new locations by computing collision-free path plans during the tree search. We evaluate both LSR variants repairing solutions from M2M and LNS-PBS over multiple 8-hour simulations across three warehouse layouts to assess long-term performance. Experimental results show that $\text{LSR}_{\text{ECR}}$ and $\text{LSR}_{\text{MCR}}$ improve overall throughput by upwards of $3.7\%$ and $5.9\%$ respectively by reducing the number of suboptimal actions by $4.4\%$ and $14.6\%$ or by $\approx10$k and $\approx36$k respectively. We additionally perform a qualitative analysis on how $\text{LSR}_{\text{ECR}}$ and $\text{LSR}_{\text{MCR}}$ reduces suboptimal actions and why they cannot in other cases.   

\section{Related Work}
\label{sec:related_work}
In this section, we situate our work within the broader context of MAPD research, which can be further broken down into two sub-problems, Multi-Robot Task Allocation (MRTA) and Multi-Agent Path Finding (MAPF). 

\subsection{Multi-Robot Task Allocation (MRTA)}

The MRTA problem answers the question of which agents will attend to what tasks in the system and when they will attend to them. A prior survey by Gerkey and Matari{\'c} \cite{gerkeysurvey} provides a useful categorization of MRTA methods depending upon if agents are capable of working on a single task (ST) or multiple tasks simultaneously (MT), if multiple robots (MR) or only a single robot (SR) can attend to a task, and if tasks are planning instantaneously (IA) or if they are time extended (TA). In this work, we focus on the ST-SR-TA problem category. Other MRTA surveys such as Korsah et al. \cite{korsahsurvey} focus on task relationship dependencies, in which we focus on the No Dependencies (ND) task category. Prior work in MRTA formulates a Traveling Salesman Problem (TSP), then uses a TSP solver \cite{liu2019task} to compute a task allocation. Other works use Large Neighborhood Search (LNS) \cite{pisinger2018large} when computing a MRTA \cite{LNS-PBS, schneider2026} or a modified version of LNS \cite{chen2021integrated}.  

\subsection{Multi-Agent Path Finding (MAPF)}

The MAPF problem has received considerable attention within and outside of MAPD. Earlier work by Silver \cite{silver2005cooperative} introduced a prioritized planning approach, Cooperative A*, which uses a prior ordering of agents in which to plan. Work by Grenouilleau et al. \cite{grenouilleau2019multi} extends Cooperative A* to their algorithm, MLA*, which plans for pairs of goal locations, which RHCR \cite{li2021lifelong} generalizes to task sequences and windowed MAPF instances. Work by Ma et al. \cite{ma2019searching} proposes Priority Based Search (PBS) which searches the space of possible priority orderings. Conflict Based Search (CBS) \cite{sharon2015conflict}, provides a complete and optimal algorithm for MAPF with the drawback of poor scalability. Barer et al. \cite{barer2014suboptimal} introduced Enhanced Conflict Based Search (ECBS), a suboptimal bounded variant of CBS, which has better scalability. More recent work has used imitation learning to learn and scale prior MAPF methods \cite{jiang2025deploying}. Recent work by Okumura \cite{okumura2023lacam} proposed LaCAM, a state-of-the-art suboptimal MAPF algorithm that reasons about constraints at agents' locations. All of these MAPF methods rely on task allocations produced by a prior algorithm, and the quality of their resulting paths is therefore directly influenced by that allocation.

\begin{figure*}[htb!]
    \centering
    \includegraphics[width=0.8\linewidth]{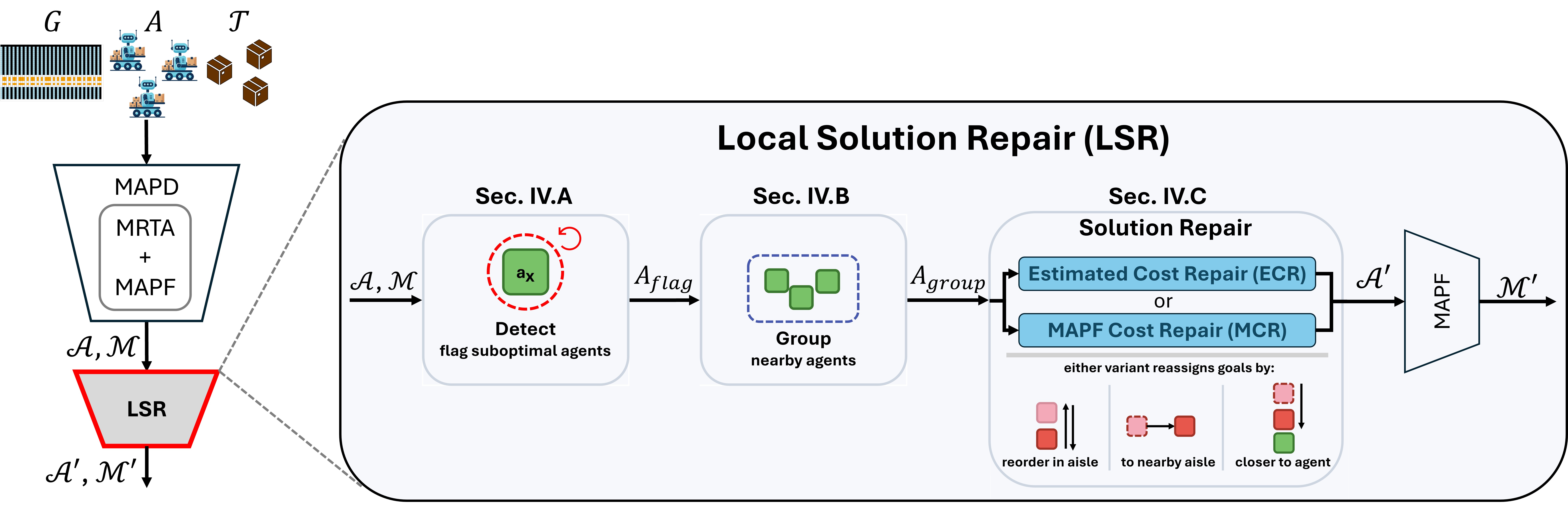}
    \caption{Overview of our Local Solution Repair (LSR) method.}
    \label{fig:system-diagram}
    \vspace{-2em}
\end{figure*}

\subsection{Multi-Agent Pickup-and-Delivery (MAPD)}

The MAPD problem can be split into two sets of algorithms, one that solves it sequentially (by solving the MRTA then MAPF problem in sequence) and algorithms that solve them simultaneously. Ma et al. \cite{ma2017lifelong} proposed the CENTRAL algorithm, which uses the Hungarian Algorithm and CBS to solve the corresponding MAPD problem. HBH-MLA* \cite{grenouilleau2019multi} uses an h-value based heuristic task allocation method with MLA* for path planning. TA-Hybrid \cite{liu2019task} models the task allocation as a special TSP variant while using an improved version of prioritized planning. MCA/RMCA \cite{chen2021integrated}, in contrast, simultaneously computes the task allocation and motion plan, which scales poorly to large MRS.

Our baselines, LNS-PBS \cite{LNS-PBS} and M2M \cite{schneider2026}, both use LNS \cite{LNS-PBS} and PBS\cite{ma2019searching} to compute a task allocation and motion plan. However, while LNS-PBS solves the MAPD problem, M2M solves the M2M-MAPD problem, an extension to MAPD which allows agents to choose the pickup and dropoff location from sets of feasible locations, requiring the task allocation to solve a 4-dimensional assignment problem. In this work, we utilize this M2M-MAPD problem to enable the system to reassign agents to new goals on-the-fly to resolve suboptimal actions within the system. 

% Our baseline method, LNS-PBS \cite{LNS-PBS}, first computes a task allocation via the Hungarian Algorithm then iteratively improves it using estimated costs via Large Neighborhood Search. LNS-PBS then computes a corresponding motion plan using the PBS MAPF algorithm. In contrast to these sequential approaches, prior work has also proposed integrated methods that solve task allocation and motion planning jointly \cite{chen2021integrated}, explicitly incorporating multi-agent path planning costs into the tasking decision. While such methods can reduce the occurrence of the suboptimal behaviors studied in this work, they incur significantly higher computational overhead due to the tight coupling between task allocation and multi-agent planning, and therefore scale poorly to large systems. \ES{add M2M}. 

\section{MAPD Problem Formulation}
\label{sec:problem_formulation}
The MAPD problem is defined with respect to $M$ agents, $A=\{a_1, a_2, ..., a_M\}$, each capable of moving through a shared environment represented as an undirected graph $G=(V, E)$ in which vertices $v_i \in V$ represent locations and edges $e_{ij} \in E$ represent traversable paths between vertices. The estimated distance to travel from $v_i \in V$ to $v_j \in V$ (i.e. shortest distance ignoring dynamic obstacles) is denoted as $d^{est}(v_i, v_j)$, while the distance returned from a Multi-Agent Path Finding (MAPF) algorithm which considers other agents is denoted as $d^{mapf}(v_i, v_j)$. The location of agent $a_i$ at timestep $t$ is denoted by $l_i(t) \in V$, and at each timestep, agents either move to an adjacent location or wait at their current location. Motion of the agent must satisfy the feasibility constraints encoded by the graph G and exclude all positions that result in collision, where a collision between agents $a_i$ and $a_j$ is defined as being co-located on the same vertex ($l_i(t)=l_j(t)$) or the same edge ($l_i(t)=l_j(t+1)\ \& \ l_i(t+1)=l_j(t)$).

We address the Many-to-Many MAPD problem \cite{schneider2026}, in which a pickup and delivery task $\tau_n\in \mathcal{T}$ is completed once an agent picks an item from one of many locations then delivers it to one of many dropoff locations. A task is defined as a tuple of two sets $\tau_n=(S_n, D_n)$, in which $S_n\subseteq V$ and $D_n \subseteq V$ are the sets of possible pickup and dropoff locations for $\tau_n$. A task $\tau_n$ allocated to agent $a_m$ is defined as $\tau_n^m=(s_p, d_q)$ where $s_p\in S_n$ and $d_q \in D_n$ are the chosen pickup and dropoff locations for task $\tau_n^m$ respectively. The current location agent $a_m$ is traveling to is referenced as a goal location $g_m\in (s_p, d_q)$, which is either the pickup or dropoff location for task $\tau_n^m$. A solution to the MAPD problem is the task allocation $\mathcal{A}$ and motion plan $\mathcal{M}$, which detail where agents will pick/place items and how the tasks are performed.

In this work, we introduce the \textbf{Suboptimality Detection and Repair (SDR) problem}: given a set of tasks $\mathcal{T}$, agents $A$, task allocation $\mathcal{A}$, and motion plan $\mathcal{M}$, compute a revised allocation and motion plan $\mathcal{A}'$ and $\mathcal{M}'$ that improves the throughput $\eta$ over the experiment by minimizing the sum of agent costs, $f: \mathcal{A} \to \mathbb{R}^+$, which sums the collision-free path cost for each agent, $f(\mathcal{A})=\sum_{a_i\in \mathcal{A}}d^{mapf}(l_i(t), g_i)$. 

\section{Local Solution Repair}
\label{sec:methods}
% \begin{figure*}[t!]
%     \centering
%     \includegraphics[width=1\linewidth]{figures/Fig2.jpg}
%     \caption{Overview of our Local Solution Repair (LSR) method.}
%     \label{fig:system-diagram}
% \end{figure*}

Below, we present how different types of suboptimal robot behavior can be detected (Section \ref{sec:detection}), grouped with other conflicting agents (Sec. \ref{sec:grouping}), then reassign the goal locations for the group of agents (Section \ref{sec:solution_repair}) through our Local Solution Repair (LSR) method (Fig. \ref{fig:system-diagram}).  
% Figure \ref{fig:system-diagram} shows an overview of our MAPD-agnostic repair approach, Local Solution Repair (LSR). 

%\SC{cut rest of paragraph if short on space, redundant with later text} During runtime, LSR detects subtoptimal behavior in task allocation ($\mathcal{A}$) and motion plan ($\mathcal{M}$), identifying a subset of agents with suboptimal plans, $A_{\text{flag}}$. For each agent $a_i \in A_{\text{flag}}$, we further identify a subset of agents in the neighborhood of $a_i$, $A^{a_i}_\text{\text{group}} \forall a_i \in A_{\text{flag}}$. Finally, we perform solution repair for each $A^{a_i}_\text{\text{group}}$ to identify improved allocations, where possible.

\subsection{Suboptimal Behavior Detection}
\label{sec:detection}

Given an existing MAPD algorithm's task allocation $\mathcal{A}$ and motion plan $\mathcal{M}$, our goal is to identify a subset of agents $A_{\text{flag}} \subset A$ which exhibit suboptimal actions. Two common suboptimal behaviors include \textbf{backtracking}, in which an agent moves away from its current goal to yield to another agent and must subsequently retrace its path, $d^{est}(l_i(t), g_i) < d^{est}(l_i(t+1), g_i)$, and \textbf{idling}, in which an agent remains stationary despite having an active goal, waiting for another agent to pass, $d^{est}(l_i(t), g_i) = d^{est}(l_i(t+1), g_i)$. To detect both the backtracking and idling behaviors, we present \textbf{Cost Estimate Deviation (CED)}. CED computes $A_{\text{flag}}$ as:

\[A_{\text{flag}}=\{a_i \in A | \frac{d^{mapf}(l_i(t), g_i) - d^{est}(l_i(t), g_i)}{d^{est}(l_i(t), g_i)} \geq \alpha  \}\]

\noindent Agent $a_i$ is added to $A_{\text{flag}}$ if its path cost exceeds its estimated shortest-path cost by $\alpha$. If an agent's goal is its current location, we cannot compute CED, and agent $a_i$ is skipped.

\subsection{Agent Grouping}
\label{sec:grouping}

Each agent $a_i \in A_{flag}$ is exhibiting suboptimal behavior to give-way to one or more other agents. Thus, our next goal is to identify the subgroup of \textit{conflicting agents} for each flagged agent to be replanned with. For each flagged agent $a_i \in A_{\text{flag}}$, we construct a group of interacting agents $A_\text{\text{group}}^{a_i}$, which includes $a_i$ and any other agents whose behavior is likely to affect its performance. The collection of all such groups is denoted $A_\text{\text{group}}$. To keep the solution repair methods in the next section computationally feasible within a 1 second time budget, we cap the size of the group of agents to  $|A_{\text{group}}^{a_i}| \le \beta$, in which if $|A_{\text{group}}^{a_i}|>\beta$, we skip repair. 

We group each flagged agent $a_i$ with agents whose goals lie in the same warehouse aisle as $g_i$. This is motivated by the aisle-dominant structure of warehouses, where narrow corridors and column-aligned goal locations induce conflicts among agents targeting the same aisle. Other grouping strategies may be more appropriate for different environments or applications.

\subsection{Solution Repair}
\label{sec:solution_repair}
We introduce two repair methods for reassigning goal locations for conflicting agents $A_{\text{group}}^{a_i}$ for $a_i\in A_{\text{flag}}$: 

\smallskip
\noindent \textbf{Estimated-Cost Repair ($\text{LSR}_{\text{ECR}}$):}
Our first method reassigns grouped agents' goal locations by solving a linear assignment problem via the Hungarian Algorithm \cite{kuhn1955hungarian} using estimated cost distances. We iterate over all $A_{\text{group}}^{a_i}$ for $a_i\in A_{\text{flag}}$ and repair each $A_{\text{group}}^{a_i}$. $\text{LSR}_{\text{ECR}}$ takes one group of agents flagged for repair ($A_{\text{group}}^{a_i}$), the current allocations of the system ($\mathcal{A}$), and the set of map vertices $V$. First, $\text{LSR}_{\text{ECR}}$ identifies a subset of candidate goal locations $L\subseteq V$ which the conflicting group of agents $A_{\text{group}}^{a_i}$ can be reassigned to. Goal locations used by other agents, $V_{alloc}$, are excluded while the possible goal locations for each agent's task $\tau^i_n$ are added. We then initialize a cost matrix $C \in \mathbb{R}^{|A^{a_i}_{\text{\text{group}}}| \times |L|}$ in which an element's cost is equal to the estimated distance from the agent's current location $l_i(t)$ to the candidate location $\ell$, otherwise, if the candidate location is not a valid goal location for $a_i$, the cost is set to $\infty$. With the cost matrix $C$, we utilize the Hungarian Algorithm to solve the linear assignment problem, assigning agents to new goal locations, minimizing total estimated distance. The Hungarian Algorithm returns a goal assignment $\sigma \in \Sigma$ which maps agents' current goal locations onto $L$, $\sigma:A_{group}^{a_i}\to L$, where $\Sigma$ is the set of possible goal assignments. We apply the goal assignment $\sigma$ onto the prior task allocation $\mathcal{A}$, returning the proposed new task assignment $\mathcal{A}'$. The algorithm will repeat this process for every agent group $A_{\text{group}}^{a_i}$ for $a_i\in A_{\text{flag}}$.

\smallskip
\noindent \textbf{MAPF-Cost Repair ($\text{LSR}_{\text{MCR}}$):}
While $\text{LSR}_{\text{ECR}}$ (above) can help resolve identified suboptimal actions, it cannot reason about conflicts between the grouped agents during goal reassignment. So, we propose our second repair method, MAPF-Cost Repair ($\text{LSR}_{\text{MCR}}$), which reassigns goals to the conflicting agent group via a Branch-and-Bound (BnB) search that computes collision-free (MAPF) path costs as it expands the tree. Like $\text{LSR}_{\text{ECR}}$, $\text{LSR}_{\text{MCR}}$ takes the same algorithm input, computes the set of possible goal locations $L \subseteq V$, and computes the cost estimate matrix $C \in \mathbb{R}^{|A^{a_i}_{\text{\text{group}}}| \times |L|}$. Next, we initialize an empty priority queue $\mathcal{Q}$, to which we append an empty solution with no allocation and $0$ cost, $n_0$. We also keep track of the best found goal assignment $\sigma$ and corresponding cost $c^*$. Once initialization is complete, we iteratively branch and prune the search tree until the priority queue is empty $\mathcal{Q}=\{\emptyset\}$, then we return the best found allocation $\sigma$. 

At the beginning of each branch-and-bound loop, we expand the node $n$ with the minimum number of remaining valid goal locations for the unassigned agents $\text{unassigned}(n)$. The lower bound cost of $n$, $c_{LB}$, is computed by summing the MAPF-informed cost for the allocated agents and the minimum estimated cost for the unallocated agents while allowing already allocated locations to ensure admissibility. \[
\begin{split}
      c_{LB} = \sum_{a_j \in \text{assigned}(n)} d^{mapf}\bigl(l_j(t), \sigma(a_j)\bigr) + \\ \sum_{a_k \in \text{unassigned}(n)} \min_{\ell \in L}C[k,\ell]
\end{split}
\]

\noindent If $c_{LB}\geq c^*$, then the best-case allocation for the remaining agents is worse than the best found solution, so $n$ is discarded. If $n$ is a complete solution and the sum-of-costs is less than the best-found solution, $c<c^*$, save $c$ as the best solution. Otherwise, we discard $n$ and pop off a new solution. Next, we select the agent $a_j$ to allocate with the minimum number of remaining feasible locations. For each remaining feasible location, $\ell \in F(a_j) \subseteq L$ for agent $a_j$, we compute $d^{mapf}(l_j(t), \ell)$ for the child node $n'$, which is then appended to $\mathcal{Q}$. We apply the final goal assignment $\sigma$ onto the prior task allocation $\mathcal{A}$, returning the proposed new task assignment $\mathcal{A}'$. We then repeat this process for every agent group $A_{group}^{a_i}\in A_{group}$. 

\subsection{Optimality}
\label{sec:optimality}
$\text{LSR}_{\text{MCR}}$ finds the minimum-cost goal assignment for $A_{\text{group}}^{a_i}$. We establish this by showing that its Branch-and-Bound search is complete and its lower bound is admissible. 
During each node expansion, a child node is generated for each valid goal location in $L$ for the selected agent, so every feasible assignment $\sigma \in \Sigma$ appears as a leaf in the search tree. The lower bound $c_{\mathrm{LB}}$ is admissible because it combines collision-aware MAPF costs for assigned agents with the minimum estimated cost for each unassigned agent, allowing repeated goals when computing the bound. Since these collision-unaware estimates lower-bound the corresponding MAPF costs, $c_{\mathrm{LB}}$ never overestimates the minimum achievable cost, allowing safe pruning while preserving the optimal in-group assignment. In contrast, $\text{LSR}_{\text{ECR}}$ is optimal only with respect to its estimated-cost objective. Because this objective does not account for inter-agent collisions, its returned assignment $A'$ is not necessarily optimal under collision-aware path costs.

% $\text{LSR}_{\text{MCR}}$ finds a minimum cost goal assignment for the in-group of agents, which we prove by showing the completeness and heuristic admissibility of our branch-and-bound search. For completeness, our expansion step for some node $n$ chooses agent $a_i$ to allocate (detailed in Sec. \ref{sec:solution_repair}). From $n$, we generate a child node for every valid goal location in $L$ to $a_i$, which trivially guarantees that there is a corresponding node in the tree for each possible agent-goal pairing while ignoring pruning. For admissibility on an arbitrary node $n$, the path cost for assigned agents $c_{mapf}$ does not consider unassigned in-group agents, so either the unassigned agents will not cause congestion (i.e. no increase in $c_{mapf}$) or they will cause delays (i.e. $c_{mapf}$ increases) as unassigned agents are assigned. When adding costs for unassigned agents from $C$, these costs ignore inter-agent collisions and locations can be selected by multiple agents, guaranteeing that they do not overestimate path costs, meaning $c_{LB}$ is admissible. So, with the completeness of the tree search and the admissible heuristic, $\text{LSR}_{\text{MCR}}$ is guaranteed to return the optimal goal assignment when considering only in-group agents. $\text{LSR}_{\text{ECR}}$ is optimal under estimated-costs; however, there is no guarantee that $\mathcal{A'}$ is the optimal goal assignment for the group of conflicting agents $A_{\text{group}}^{a_i}$, as its estimated cost function $d^{est}(v_i,v_j)$ does not consider inter-agent collisions. 

More generally, neither LSR$_{\text{MCR}}$ nor LSR$_{\text{ECR}}$ guarantees that a proposed local reassignment improves the global solution, because neither accounts for interactions between in-group and out-of-group agents. For example, reassigning an agent to a nearby aisle may introduce congestion with agents outside the repaired group. This motivates the solution acceptance criterion described next.

%When considering both in-group and out-of-group agents, $\text{LSR}_{\text{MCR}}$ and $\text{LSR}_{\text{ECR}}$ cannot guarantee that the proposed allocation is better than the prior. This is due to both methods not considering collisions between agents inside and outside the group (e.g., an agent's goal reassigned to a nearby aisle causes congestion with outside-agent(s) within the aisle).

% result in a equal or better quality global solution because neither method considers the changes to outside-group agent paths from the goal changes for in-group agents (e.g., if an agent is reassigned to a nearby aisle, in which some number of outside-group agents are entering or exiting, the addition of the in-group agent results in increased congestion which neither method considers). 

\subsection{Solution Acceptance Criterion}
\label{sec:acceptance_criterion}
When $\text{LSR}_{\text{ECR}}$ or $\text{LSR}_{\text{MCR}}$ returns the proposed task allocation $\mathcal{A'}$, we must decide whether to accept or reject it. The proposed task allocation $\mathcal{A'}$ is accepted if and only if the sum-of-path-costs is less than the prior allocation $\mathcal{A'}$,   $f(\mathcal{A}')<f(\mathcal{A})$. Otherwise $\mathcal{A'}$ is rejected and $\mathcal{A}$ is kept. This ensures the returned allocation is always the same or better than the prior allocation.

\section{Experiments}
\label{sec:experiments}
In this section, we provide an overview of our experimental design, experimental conditions, experiment and method parameters, and metrics measured during experiments, to assess the overall benefit of LSR. 

\begin{figure}
    \centering
    \includegraphics[width=0.9\linewidth]{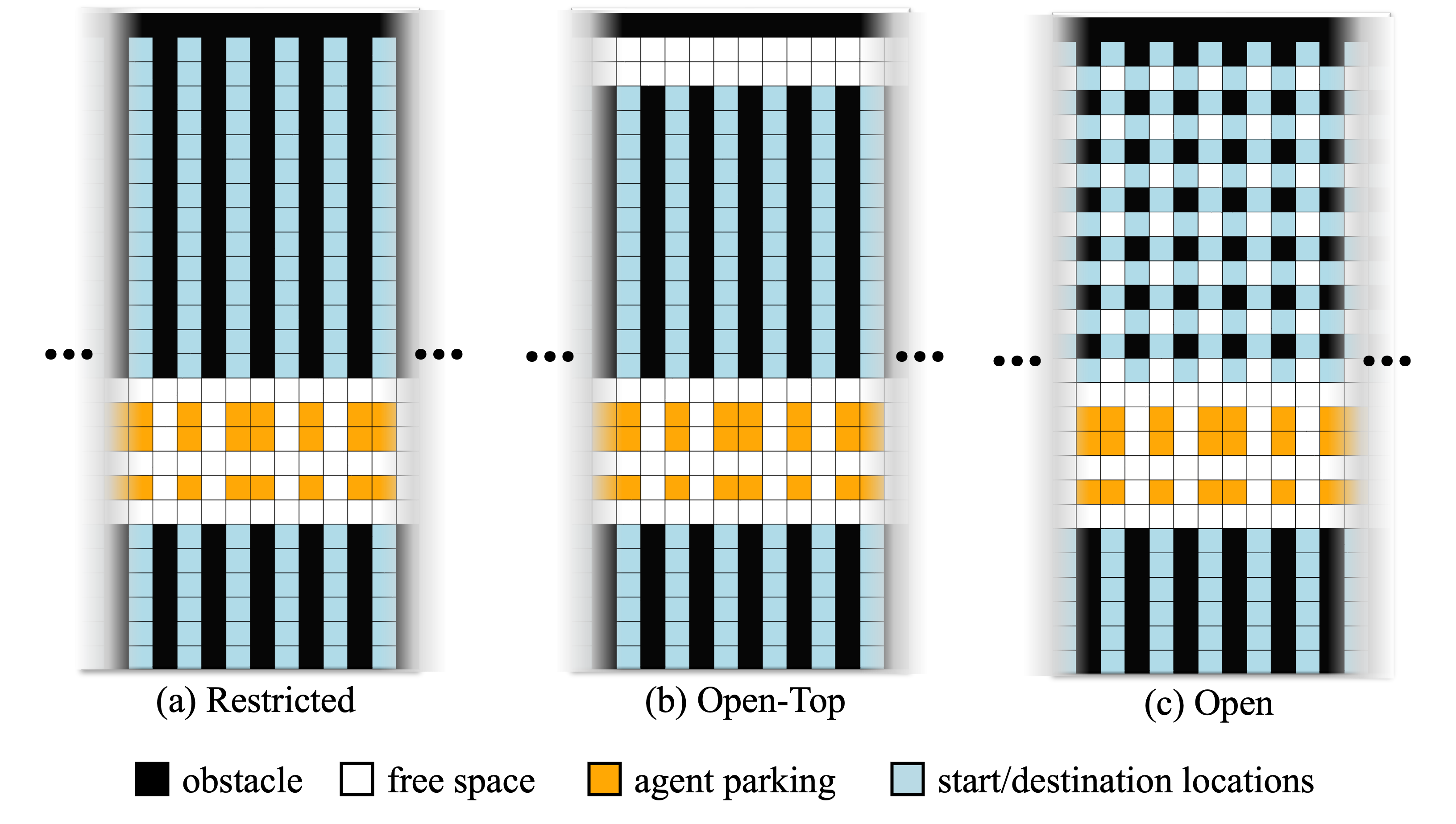}
    \caption{Partial view of our three simulated warehouse layouts. All non-obstacle regions are traversable by agents, subject to collision constraints. Each map is 27$\times$ 50.}
    \label{fig:maps}
    \vspace{-1em}
\end{figure}

\smallskip
\noindent \textit{Experimental Setup:} We evaluate the performance of our methods in a simulated warehouse domain, which is designed to capture the complex nature of real-world MAPD problems. We evaluate our methods across three different 2D warehouse layouts shown in Figure \ref{fig:maps}. The restricted map contains long aisles with a single entry point close to the item inbound/outbound loading area at the bottom. The open-top map has additional open space at the top of the map, allowing agents to enter or exit aisles at either end. The open map contains a grid layout for the inventory. These map layouts were modeled after similar maps found in prior MAPD and MAPF works \cite{li2021lifelong, liu2019task}. In all these environments, items enter and exit the warehouse at the bottom of the map, while items are stored and retrieved in the top portion. All three maps mimic layouts currently in use in warehouse logistics operations.  

\smallskip
\noindent \textit{Experimental Conditions:} To evaluate the benefit of LSR, we compare the performance of the following three methods: 

\begin{flushitemize}
    \item \textbf{M2M:} M2M \cite{schneider2026} is a state-of-the-art MAPD approach which relaxes the assumption that each task contains a predefined pickup and dropoff location and allows each task to be completed by visiting a pickup and dropoff location from a set of acceptable pickup and dropoff locations.
    \item \textbf{LNS-PBS:} As discussed in Section \ref{sec:related_work}, LNS-PBS \cite{LNS-PBS} is a prior MAPD approach that is shown to be highly effective and computationally efficient; however, this method is designed to solve \textit{one-to-one} MAPD instances with only one option for pickup and dropoff $\tau_n=(s_p, d_q)$. Prior to running LNS-PBS, we select $s_p \in S_n$ and $d_q \in D_n$ such that each task is assigned the closest pair of pickup and dropoff locations ($\arg\min_{\substack{s_p \in S_n, d_q \in D_n}} \; d^{est}(s_p, d_q)$).
    \item \textbf{$\text{LSR}_{\text{ECR}}$:} Estimated Cost Repair ($\text{LSR}_{\text{ECR}}$) method applied both baselines: \textbf{LNS-PBS + $\text{LSR}_{\text{ECR}}$} and \textbf{M2M + $\text{LSR}_{\text{ECR}}$}.
    \item \textbf{$\text{LSR}_{\text{MCR}}$:} MAPF-Cost Repair ($\text{LSR}_{\text{MCR}}$) method applied to both baselines: \textbf{LNS-PBS + $\text{LSR}_{\text{MCR}}$} and \textbf{M2M + $\text{LSR}_{\text{MCR}}$}.
\end{flushitemize}

\begin{figure*}[htbp]
    \centering
    % Row 1
    \begin{subfigure}{0.3\textwidth}
        \centering
        \includegraphics[width=\textwidth]{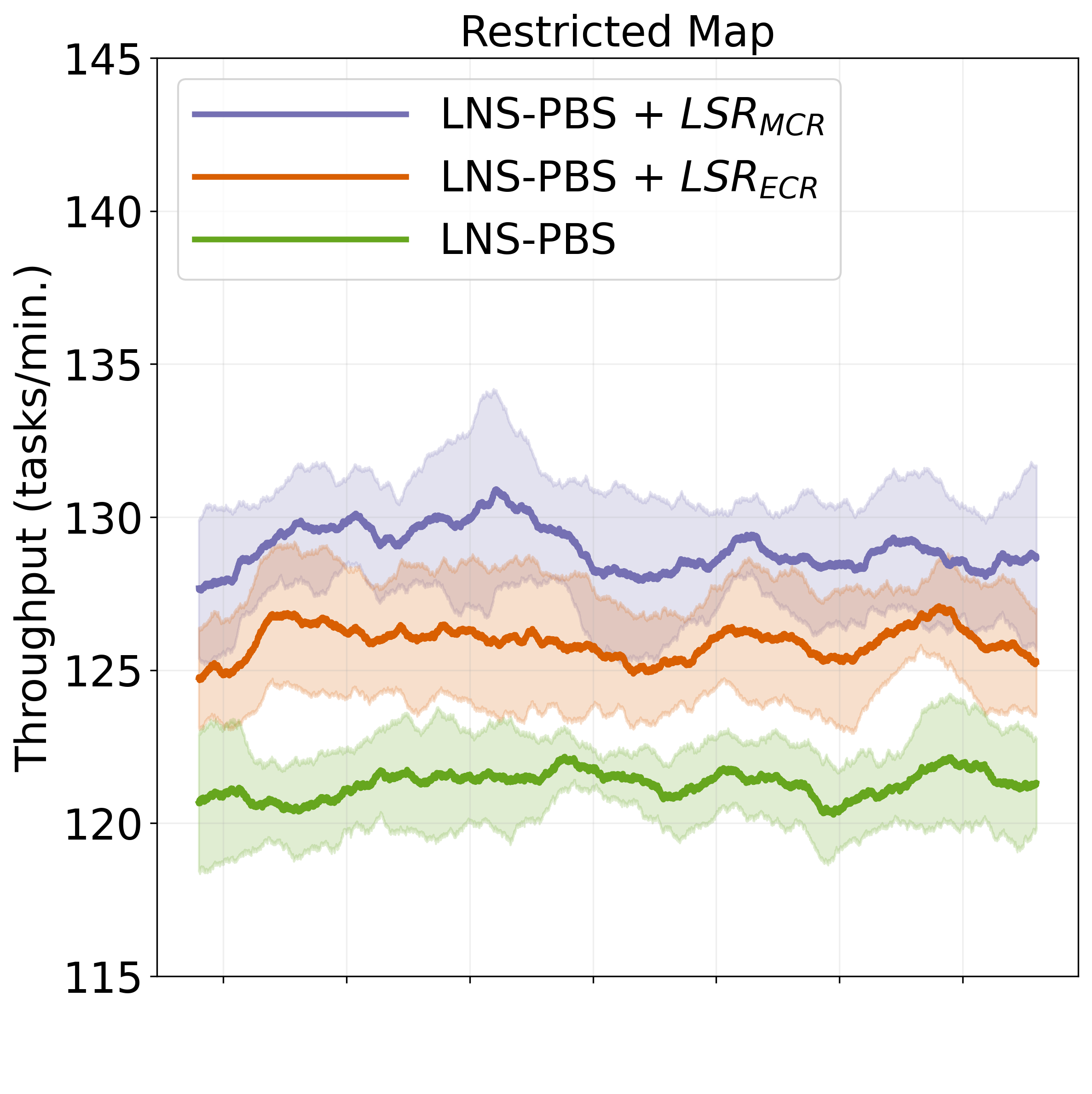}
        \label{fig:1}
    \end{subfigure}
    \hspace{-0.80em}
    \begin{subfigure}{0.3\textwidth}
        \centering
        \includegraphics[width=\textwidth]{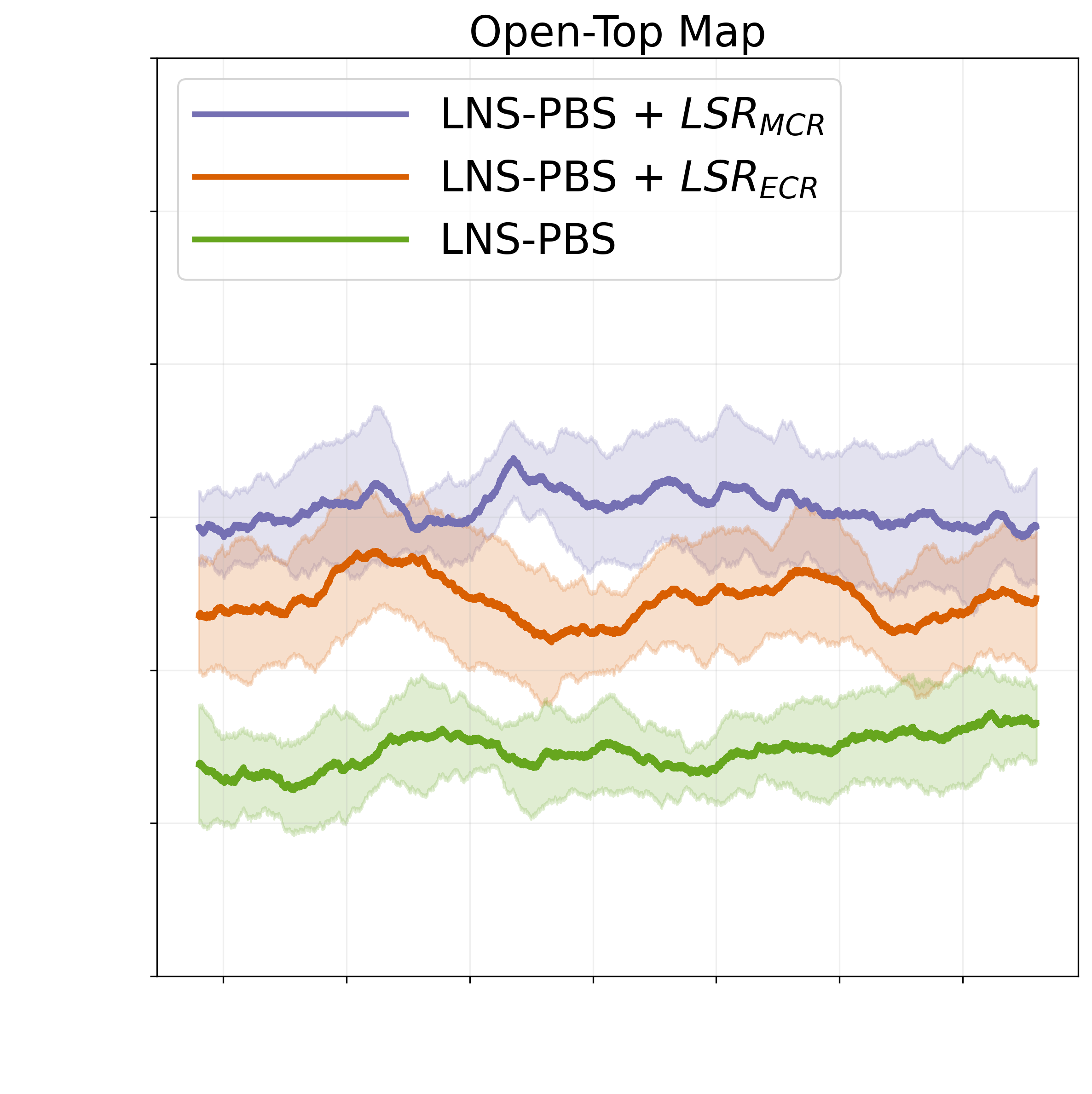}
        \label{fig:2}
    \end{subfigure}
    \hspace{-0.80em}
    \begin{subfigure}{0.3\textwidth}
        \centering
        \includegraphics[width=\textwidth]{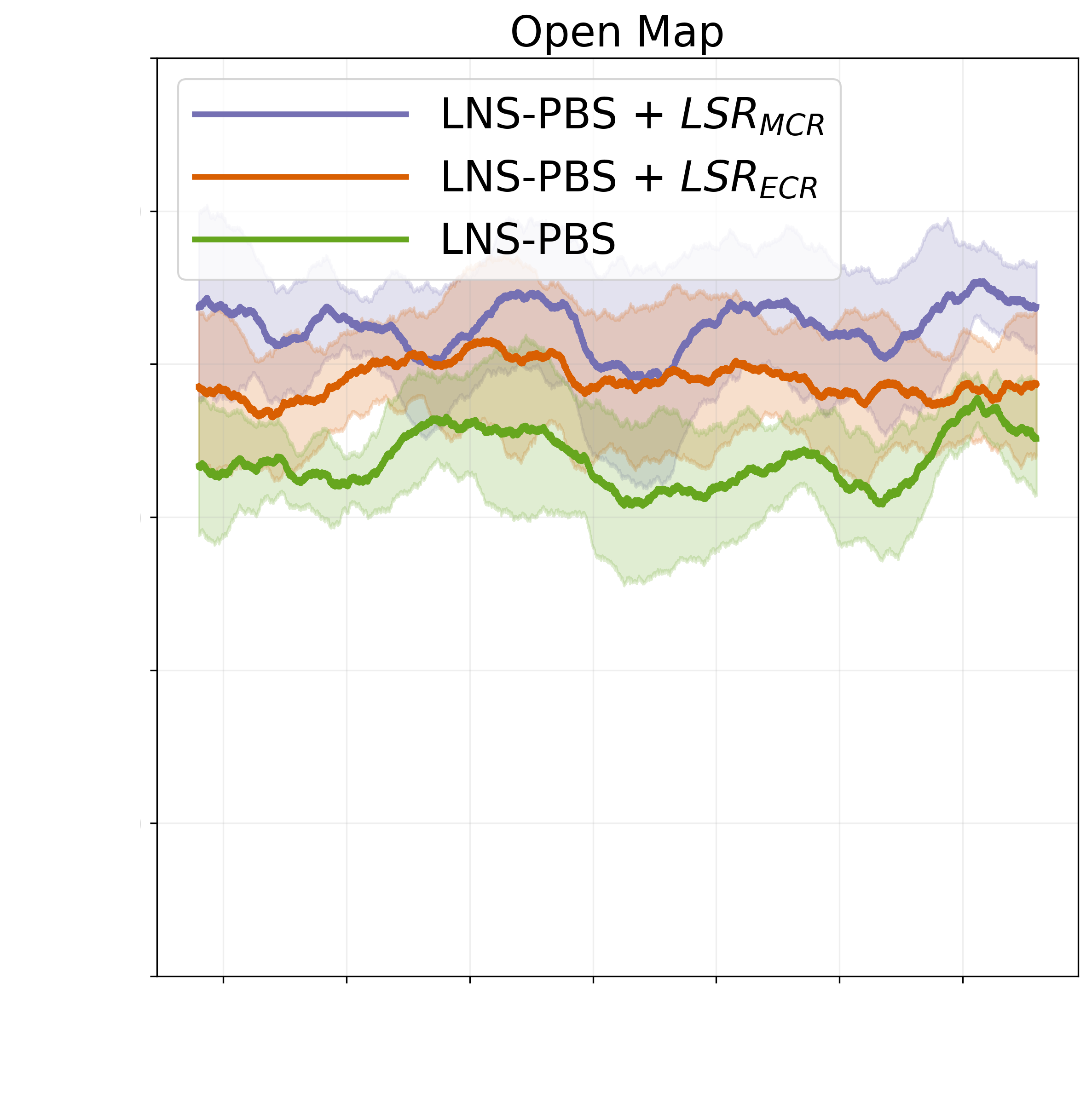}
        \label{fig:3}
    \end{subfigure}

    \vspace{-2.5em} % vertical gap between rows

    % Row 2
    \begin{subfigure}{0.3\textwidth}
        \centering
        \includegraphics[width=\textwidth]{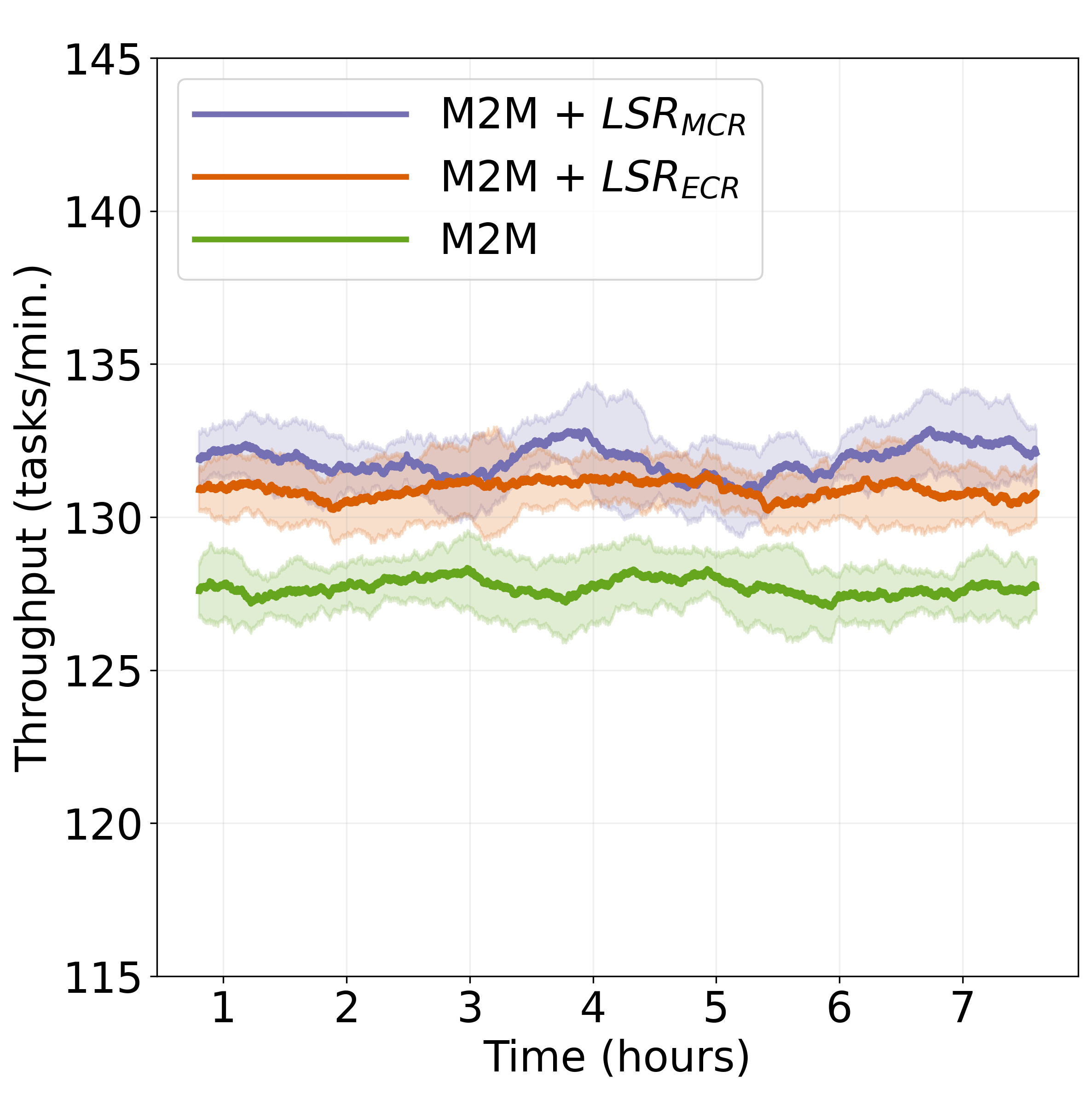}
        \label{fig:4}
    \end{subfigure}
    \hspace{-0.80em}
    \begin{subfigure}{0.3\textwidth}
        \centering
        \includegraphics[width=\textwidth]{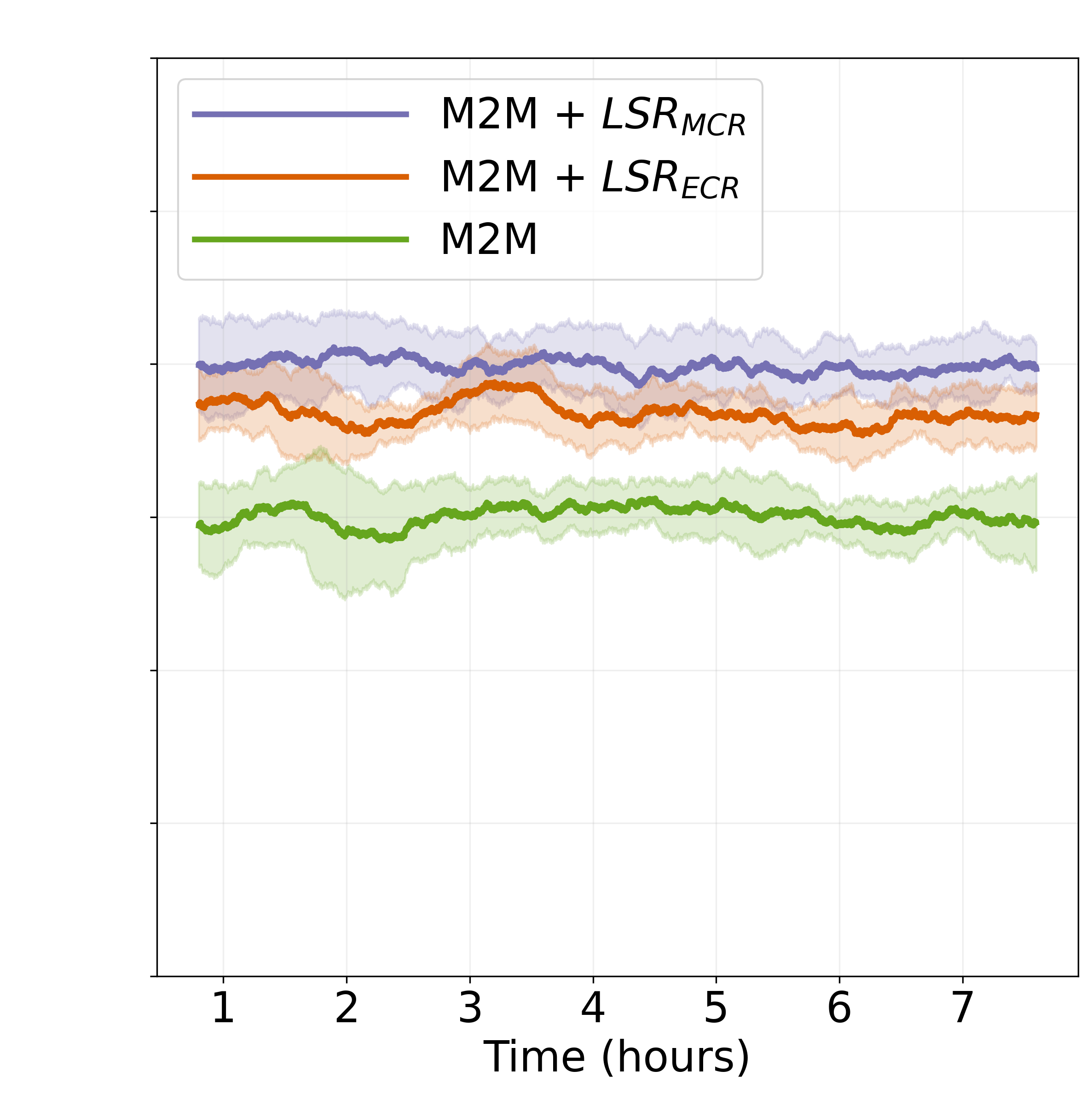}
        \label{fig:5}
    \end{subfigure}
    \hspace{-0.80em}
    \begin{subfigure}{0.3\textwidth}
        \centering
        \includegraphics[width=\textwidth]{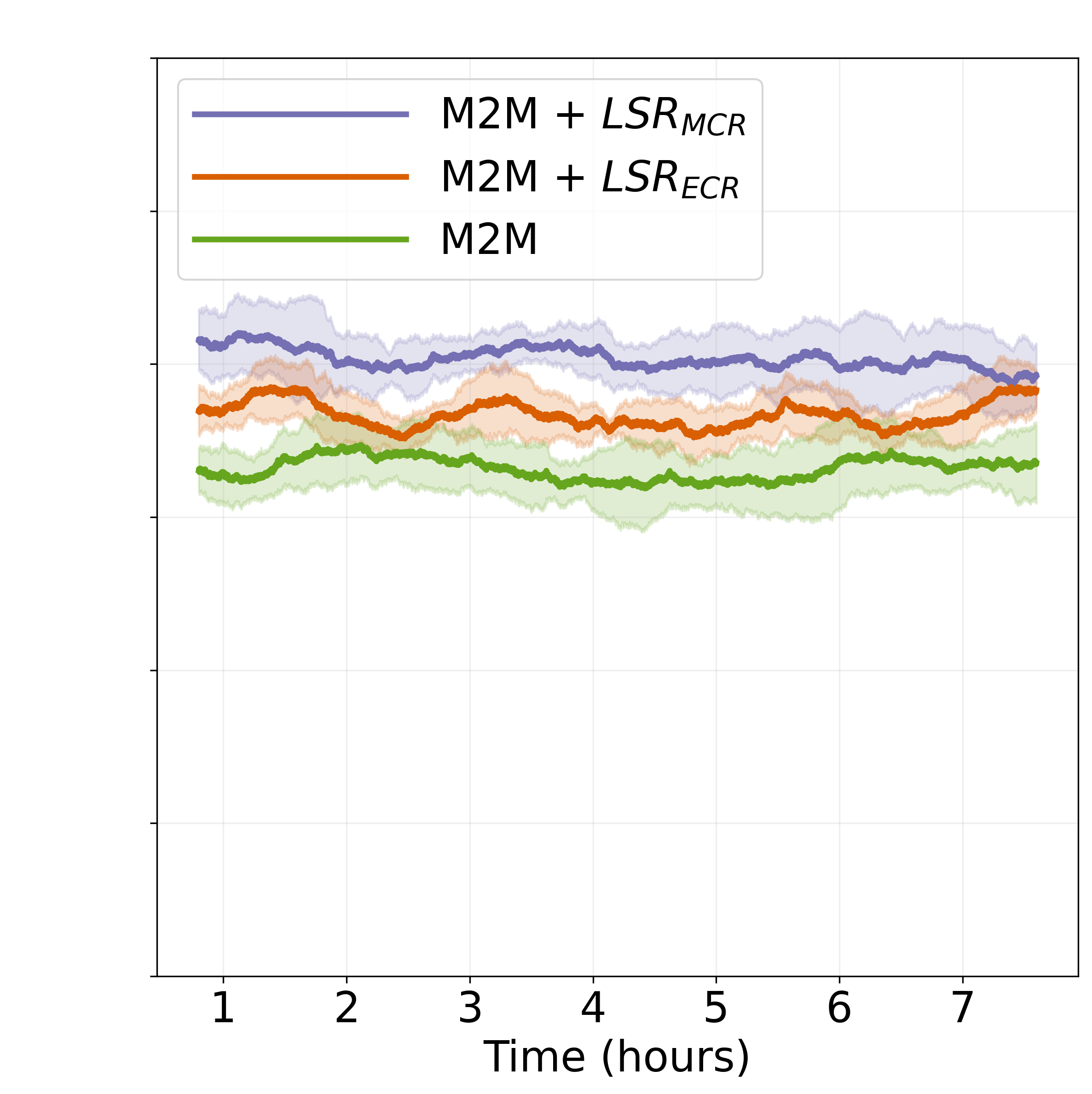}
        \label{fig:6}
    \end{subfigure}

    \vspace{-1em}
    \caption{Comparison of the rolling mean (line) and std-dev (shaded region) of $\eta$ (\textbf{$\uparrow$ is better}) reported across Restricted (a), Open-Top (b), and Open (c) maps. Our repair methods $\text{LSR}_{\text{ECR}}$ (orange) and $\text{LSR}_{\text{MCR}}$ (purple) consistently outperform the baselines (green), LNS-PBS (top row) and M2M (bottom row) across all map types.}
    \label{fig:TH}
\end{figure*}

\begin{table*}[h]
\centering
\begin{tabular}{|c|ccc|ccc|ccc|}
\hline
Method                              & \multicolumn{3}{c|}{Restricted}                                                                                                                                                          & \multicolumn{3}{c|}{Open-Top}                                                                                                                                                              & \multicolumn{3}{c|}{Open}                                                                                                                                                                \\ \hline
Metrics
                                    & \multicolumn{1}{c|}{$\eta$ (tasks/min.)}                                          & \multicolumn{1}{c|}{$\Delta_{\eta}\%$}                                & \multicolumn{1}{l|}{$R_{\text{subopt}}\%$} & \multicolumn{1}{c|}{$\eta$ (tasks/min.)}                                            & \multicolumn{1}{c|}{$\Delta_{\eta}\%$}                                & \multicolumn{1}{l|}{$R_{\text{subopt}}\%$} & \multicolumn{1}{c|}{$\eta$ (tasks/min.)}                                          & \multicolumn{1}{c|}{$\Delta_{\eta}\%$}                                & \multicolumn{1}{l|}{$R_{\text{subopt}}\%$} \\ \hline
\rowcolor[HTML]{E3E2EF} 
LNS-PBS + $\text{LSR}_{\text{MCR}}$ & \multicolumn{1}{c|}{\cellcolor[HTML]{E3E2EF}\textbf{$128.8\pm 0.6$}} & \multicolumn{1}{c|}{\cellcolor[HTML]{E3E2EF}\textbf{$+5.9\%$}} & \textbf{$+14.4\%$}                                & \multicolumn{1}{c|}{\cellcolor[HTML]{E3E2EF}\textbf{$130.1\pm 0.7$}} & \multicolumn{1}{c|}{\cellcolor[HTML]{E3E2EF}\textbf{$+5.9\%$}} & \textbf{$+14.6\%$}                                & \multicolumn{1}{c|}{\cellcolor[HTML]{E3E2EF}\textbf{$136.1\pm 0.7$}} & \multicolumn{1}{c|}{\cellcolor[HTML]{E3E2EF}\textbf{$+3.2\%$}} & \textbf{$+12.1\%$}                                \\ \hline
\rowcolor[HTML]{F7DFCC} 
LNS-PBS + $\text{LSR}_{\text{ECR}}$ & \multicolumn{1}{c|}{\cellcolor[HTML]{F7DFCC}$125.8\pm 0.6$}          & \multicolumn{1}{c|}{\cellcolor[HTML]{F7DFCC}$+3.7\%$}          & $+4.1\%$                                         & \multicolumn{1}{c|}{\cellcolor[HTML]{F7DFCC}$127.6\pm 0.4$}            & \multicolumn{1}{c|}{\cellcolor[HTML]{F7DFCC}$+4.1\%$}          & $+4.2\%$                                          & \multicolumn{1}{c|}{\cellcolor[HTML]{F7DFCC}$134.5\pm 0.8$}          & \multicolumn{1}{c|}{\cellcolor[HTML]{F7DFCC}$+2.0\%$}          & $+4.0\%$                                          \\ \hline
\rowcolor[HTML]{9B9B9B} 
\cellcolor[HTML]{E0EDD2}LNS-PBS     & \multicolumn{1}{c|}{\cellcolor[HTML]{E0EDD2}$121.2\pm 0.4$}          & \multicolumn{1}{c|}{\cellcolor[HTML]{9B9B9B}}                  &                                                  & \multicolumn{1}{c|}{\cellcolor[HTML]{E0EDD2}$122.4\pm 0.4$}            & \multicolumn{1}{c|}{\cellcolor[HTML]{9B9B9B}}                  &                                                  & \multicolumn{1}{c|}{\cellcolor[HTML]{E0EDD2}$131.8\pm 0.6$}          & \multicolumn{1}{c|}{\cellcolor[HTML]{9B9B9B}}                  &                                                  \\ \hline
\rowcolor[HTML]{E3E2EF} 
M2M + $\text{LSR}_{\text{MCR}}$     & \multicolumn{1}{c|}{\cellcolor[HTML]{E3E2EF}\textbf{$131.8\pm 0.4$}} & \multicolumn{1}{c|}{\cellcolor[HTML]{E3E2EF}\textbf{$+3.2\%$}} & \textbf{$+12.7\%$}                                & \multicolumn{1}{c|}{\cellcolor[HTML]{E3E2EF}\textbf{$134.9\pm 0.3$}}   & \multicolumn{1}{c|}{\cellcolor[HTML]{E3E2EF}\textbf{$+3.7\%$}} & \textbf{$+13.8\%$}                                & \multicolumn{1}{c|}{\cellcolor[HTML]{E3E2EF}\textbf{$135.1\pm 0.6$}} & \multicolumn{1}{c|}{\cellcolor[HTML]{E3E2EF}\textbf{$+2.6\%$}} & \textbf{$+10.3\%$}                                \\ \hline
\rowcolor[HTML]{F7DFCC} 
M2M + $\text{LSR}_{\text{ECR}}$     & \multicolumn{1}{c|}{\cellcolor[HTML]{F7DFCC}$130.8\pm 0.2$}          & \multicolumn{1}{c|}{\cellcolor[HTML]{F7DFCC}$+2.4\%$}          & $+4.4\%$                                          & \multicolumn{1}{c|}{\cellcolor[HTML]{F7DFCC}$133.2\pm 0.3$}            & \multicolumn{1}{c|}{\cellcolor[HTML]{F7DFCC}$+2.5\%$}          & $+4.2\%$                                          & \multicolumn{1}{c|}{\cellcolor[HTML]{F7DFCC}$133.8\pm 0.2$}          & \multicolumn{1}{c|}{\cellcolor[HTML]{F7DFCC}$+1.6\%$}          & $+4.2\%$                                          \\ \hline
\rowcolor[HTML]{9B9B9B} 
\cellcolor[HTML]{E0EDD2}M2M         & \multicolumn{1}{c|}{\cellcolor[HTML]{E0EDD2}$127.6\pm 0.3$}          & \multicolumn{1}{c|}{\cellcolor[HTML]{9B9B9B}}                  &                                                  & \multicolumn{1}{c|}{\cellcolor[HTML]{E0EDD2}$129.9\pm 0.5$}            & \multicolumn{1}{c|}{\cellcolor[HTML]{9B9B9B}}                  &                                                  & \multicolumn{1}{c|}{\cellcolor[HTML]{E0EDD2}$131.6\pm 0.3$}          & \multicolumn{1}{c|}{\cellcolor[HTML]{9B9B9B}}                  &                                                  \\ \hline
\end{tabular}%
\caption{Throughput ($\eta$), percent change in throughput ($\Delta_{\eta}\%$), and percent change in suboptimal actions ($R_{\text{subopt}}\%$) across baseline methods, map types, and repair methods, averaged over 10 simulations each. $\text{LSR}_{\text{MCR}}$ performs best, reducing suboptimal actions by $14.6\%$, increasing throughput by $5.9\%$.}
\label{tab:TH}
\vspace{-2em}
\end{table*}

\smallskip
\noindent \textit{Parameters:}
We perform an ablation for LSR and set the hyperparameter $\alpha=0.1$ for our suboptimality detection threshold. We set $\beta=3$ for $\text{LSR}_{\text{MCR}}$ and $\beta=4$ for $\text{LSR}_{\text{ECR}}$. We set the hyperparameters for LNS-PBS the same as in Xu et al. \cite{LNS-PBS}. For our simulated environment, our maps were 27$\times$50, the number of agents was set to $M=40$. Every timestep, 4 tasks were released into the system with a limit of 120 total active tasks. We had 30 unique item types in the warehouse with an initial uniformly distributed inventory layout at 25\% warehouse capacity. The performance for each method is reported as the average of 10 warehouse simulations, in which each simulation lasted for 8 simulated hours. LSR\footnote{https://github.com/Ethan-Schneider/LSR.git} and the repair methods were implemented in Python, while the implementation of LNS-PBS was provided by the original authors in C++. 

\smallskip
\noindent \textit{Metrics:} We evaluate the performance between our experimental conditions with the following metrics:

\begin{flushitemize}
    \item \textbf{Throughput ($\eta$):} calculated as the number of tasks completed per minute. We report the mean, median, and standard deviation throughput over multiple simulations and maps. 
    \item \textbf{Percent Improvement of Throughput ($\Delta_{\eta}\%$):} calculated as the percent improvement of $\eta$ of the repair method over the baseline.
    \item \textbf{Percent Reduction in Suboptimal Actions ($R_{\text{subopt}}\%$):} calculated as the percentage decrease in suboptimal actions achieved by the repair method relative to the baseline.
    % \item \textbf{Idling and Backtracking Instances:} reported as the number of idling actions, backtracking actions, and total suboptimal actions averaged over the multiple simulations per map type, baseline, and repair method.
    \item \textbf{Repair Mechanisms:} reported as codes identified from visualized accepted repair scenarios. 
    \item \textbf{Repair Rejections:} reported as codes identified from visualized rejected repair scenarios. 
    \item \textbf{Repair Computation Time:} reported by the mean and standard deviation computation time, in seconds, for each repair cycle, averaged over multiple simulations and environments. 
\end{flushitemize}

\section{Results}
\label{results}
In this section, we report results for each of our metrics.

\subsection{Performance Gains from Repair}
We first evaluate whether LSR improves overall system throughput and how these gains relate to reductions in suboptimal agent behavior.
As shown in Figure \ref{fig:TH} and Table \ref{tab:TH}, both LSR repair methods improve throughput $\eta$ across all baseline methods and map types.  $\text{LSR}_{\text{MCR}}$ yields the largest gains, increasing $\eta$ by up to $5.9\%$ for LNS-PBS and $3.7\%$ for M2M. Across baseline methods, LSR produces larger improvements when applied to LNS-PBS than to M2M. We attribute this difference to M2M's ability to reason over multiple possible pickup and dropoff locations during its initial task allocation, whereas LNS-PBS selects these locations before task allocation. LSR therefore provides LNS-PBS with greater opportunity to improve the initial goal assignments. Across map types, the Restricted and Open-Top maps show larger gains than the Open map, where the grid structure provides agents with more opportunities to navigate around one another and reduces the impact of local conflicts.

To understand what drives the throughput gains, we examine the reduction in suboptimal actions ($R_{\text{subopt}}\%$) that results from LSR.
$\text{LSR}_{\text{MCR}}$ reduces suboptimal actions by $14.6\%$, compared with a $4.4\%$ reduction for  $\text{LSR}_{\text{ECR}}$.  Consistent with the throughput results, the greatest reduction in suboptimal actions occurs in the Restricted and Open-Top maps due to their narrow aisles. Whereas, in the Open map, a fewer percentage of suboptimal actions are corrected by either repair method. 

% Our results show that $\text{LSR}_{\text{MCR}}$ produces improved repair solutions $2.5-3.5\times$ more frequently than $\text{LSR}_{\text{ECR}}$.

% %LSR reduces suboptimal actions taken by agents across every baseline and map type.  
%  $\text{LSR}_{\text{MCR}}$ reduces over $36$k suboptimal actions, equivalent to $14.6\%$ fewer suboptimal actions on average, compared to a $10$k reduction, $4.4\%$ reduction, by $\text{LSR}_{\text{ECR}}$.

\textbf{Summary:} Both $\text{LSR}_{\text{MCR}}$ and $\text{LSR}_{\text{ECR}}$ significantly improve the overall throughput of LNS-PBS and M2M across all map types. The greatest improvement is shown by LNS-PBS + $\text{LSR}_{\text{MCR}}$ in the Restricted and Open-Top maps, which results in a $5.9\%$ percent increase in $\eta$, or $128.8$ (tasks/min.) and $130.1$ (tasks/min.) for the two maps. This increase in $\eta$ is a result of LNS-PBS + $\text{LSR}_{\text{MCR}}$ decreasing suboptimal actions by $14.4\%$ and $14.6\%$ within the two maps. 

\subsection{Repair Outcomes and Mechanisms}
\label{sec:repair-analysis}

Above, we showed that LSR improves throughput by identifying and repairing suboptimal behavior. In this section, we examine how these repairs function. We first measure how often a proposed goal reassignment improves the current solution and is accepted. We then analyze \textit{how} accepted repairs improve the solution and \textit{why} rejected repairs fail.

\begin{figure}[h]
    \centering
    \includegraphics[width=0.9\linewidth]{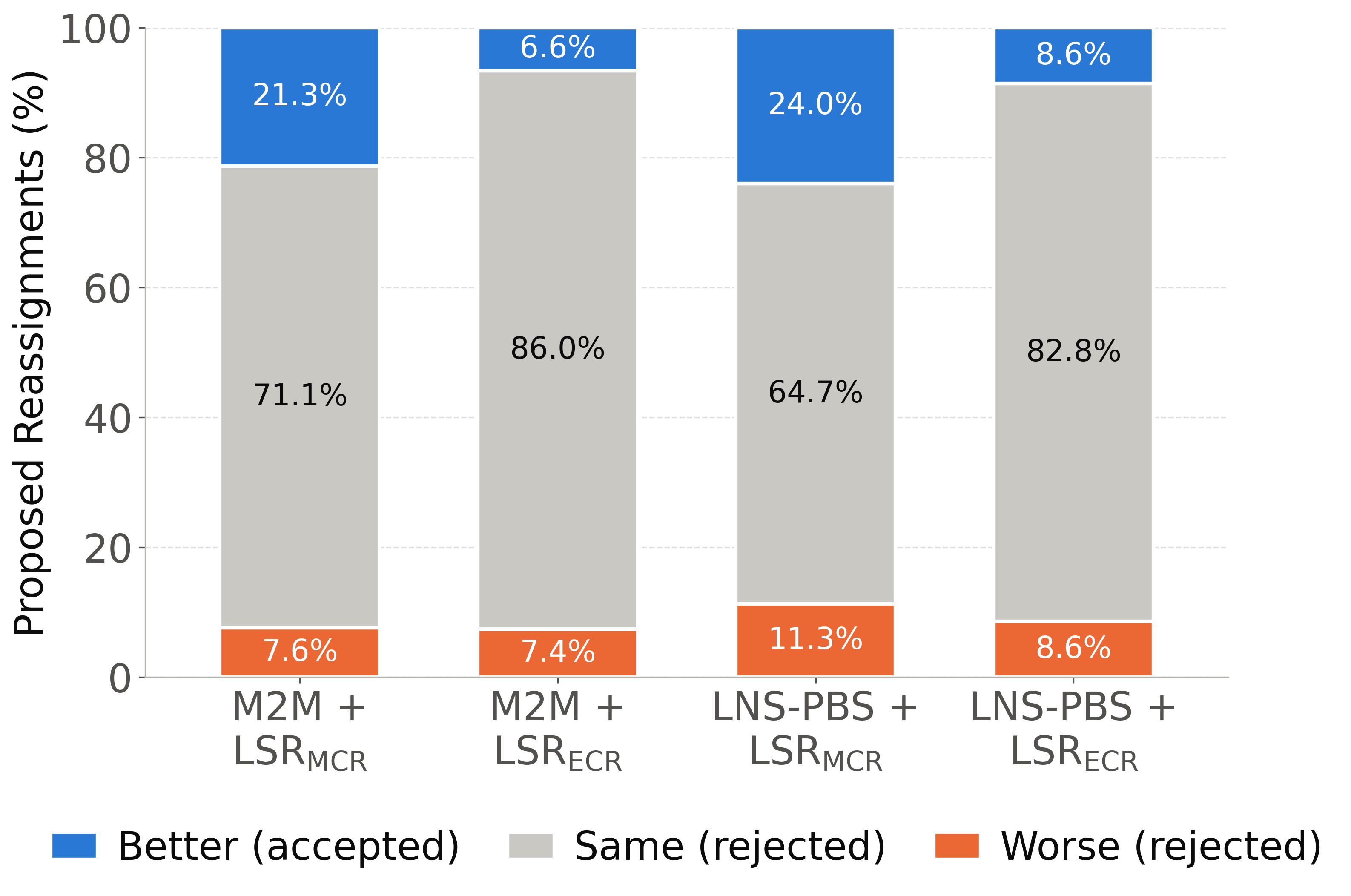}
    \caption{Breakdown of better (accepted), same (rejected), and worse (rejected) per combination of baseline method and repair method averaged across map types and seeds. $\text{LSR}_{\text{MCR}}$ proposes $\approx 3\times$ more goal reassignments that are accepted than $\text{LSR}_{\text{ECR}}$. Across baseline methods, LNS-PBS has $\approx 2\%$ more accepted goal reassignments than M2M.}
    \label{fig:proposed-reassignment-breakdown}
    \vspace{-1em}
\end{figure}

\smallskip
\noindent\textbf{Repair Acceptance} After suboptimal behavior is detected and the corresponding agents are grouped, LSR proposes a new goal assignment (Sec.~\ref{sec:solution_repair}). Figure~\ref{fig:proposed-reassignment-breakdown} reports the proportion of proposed reassignments that result in better, equivalent, or worse solutions. Across both baseline methods, $\text{LSR}_{\text{MCR}}$ has an acceptance rate of over $21\%$, compared to just $6.6\%$ and $8.6\%$ for $\text{LSR}_{\text{ECR}}$. This difference reflects the cost functions used during reassignment. $\text{LSR}_{\text{ECR}}$ relies on estimated, collision-unaware distances ($d^{est}$), whereas $\text{LSR}_{\text{MCR}}$ uses collision-aware MAPF costs
($d^{mapf}$). As a result, $\text{LSR}_{\text{MCR}}$ can account for conflicts among grouped agents when selecting new goals, producing useful reassignments more frequently.

\begin{figure}[t]
    \centering
    \includegraphics[width=0.9\linewidth]{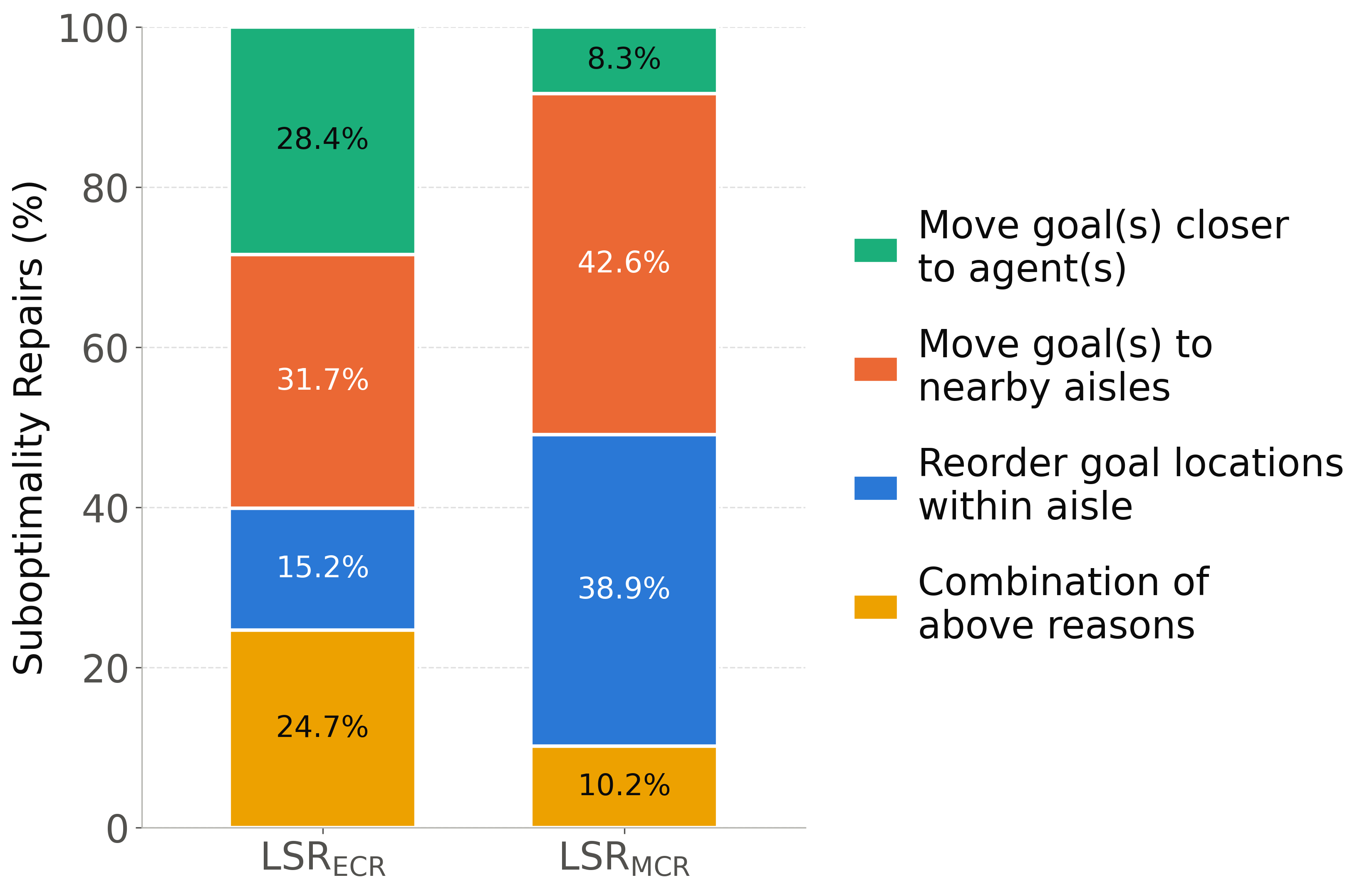}
    \caption{Distribution of repair mechanisms for $\text{LSR}_{\text{ECR}}$ and $\text{LSR}_{\text{MCR}}$ ($n=300$ each) pooled across baselines and map types. $\text{LSR}_{\text{MCR}}$'s conflict-aware costs let it reason about goal order within aisle and goals between aisles, whereas $\text{LSR}_{\text{ECR}}$ relies on inter-aisle and closer goal location goal reassignments through collision-unaware shortest-path distances.}
    \label{fig:suboptimality_improvements}
    \vspace{-1em}
\end{figure}

\smallskip
\noindent\textbf{Accepted Repair Mechanisms} To understand how accepted repairs improve solution quality, we analyzed $600$ randomly selected successful repairs ($n=300$ per repair method) and classified the resulting solutions into four mechanisms: (1) reorder goal locations within an aisle, (2) move one or more goals to a nearby aisle, (3) move goals closer to agents' current locations, or (4) a combination of these mechanisms. Figure~\ref{fig:suboptimality_improvements} shows their distribution.

We observe that the two methods produce improvements through different mechanisms. $\text{LSR}_{\text{MCR}}$, which reasons about inter-agent conflicts during goal reassignment, primarily repairs solutions by moving goals to nearby aisles ($42.6\%$) or reordering goals within an aisle ($38.9\%$). The remaining repairs move goals closer to agents ($8.3\%$) or combine multiple mechanisms ($10.2\%$). In contrast, $\text{LSR}_{\text{ECR}}$ assigns goals using collision-unaware distance and therefore relies more heavily on moving goals to nearby aisles ($31.7\%$), moving goals closer to agents ($28.4\%$), or combinations of these mechanisms ($24.7\%$). It also reorders goals within an aisle in $15.2\%$ of accepted repairs, but these changes result from relative estimated distances rather than explicit reasoning about within-aisle conflicts.

\begin{figure}[t]
    \centering
    \includegraphics[width=0.9\linewidth]{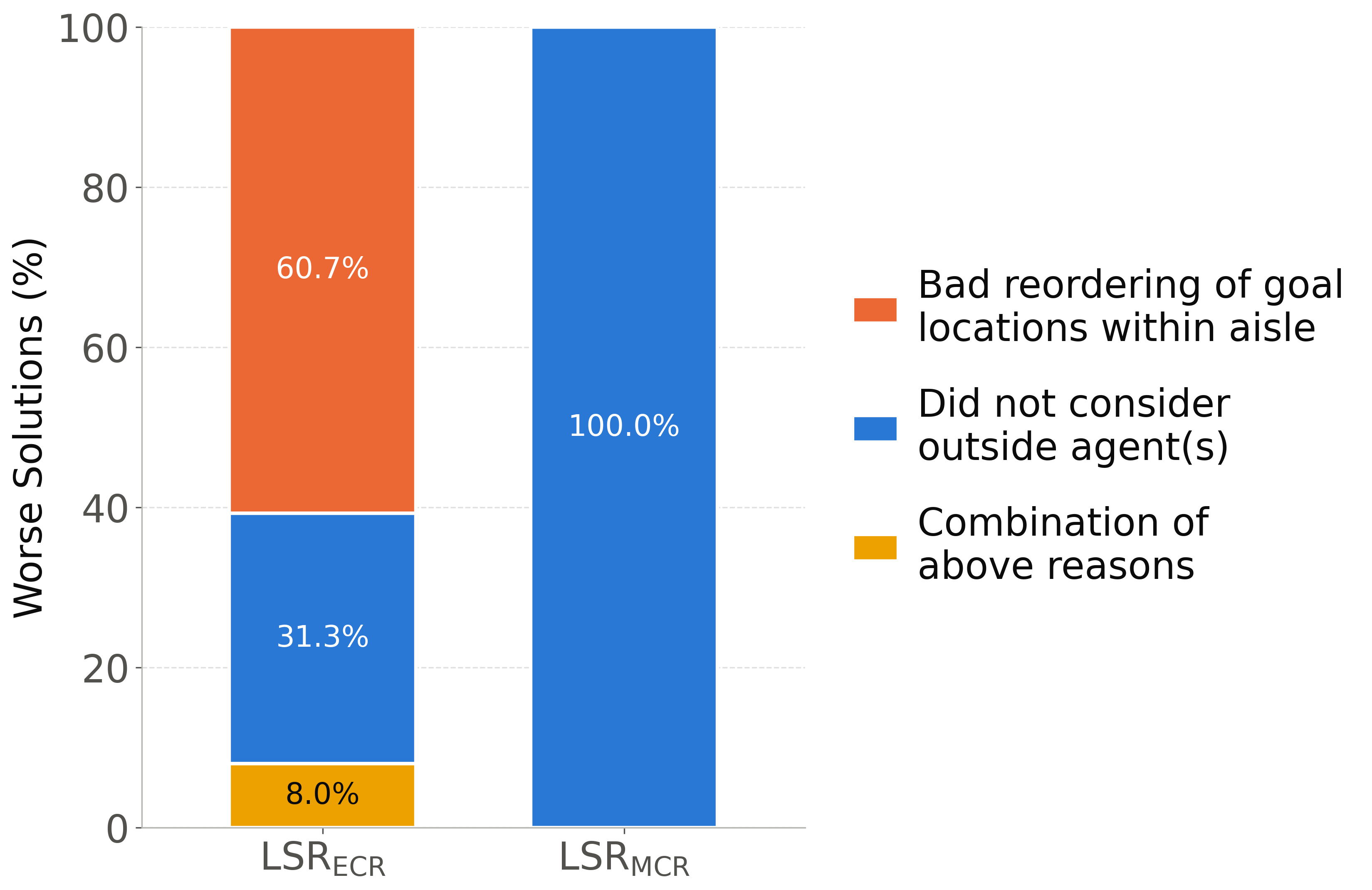}
    \caption{Distribution of worse (rejected) goal reassignments for $\text{LSR}_{\text{ECR}}$ and $\text{LSR}_{\text{MCR}}$ ($n=150$ each) pooled across baselines and map types. $\text{LSR}_{\text{MCR}}$ suggests worse goal reassignments due to goal changes affecting outside-group agents. $\text{LSR}_{\text{ECR}}$ suggests worse solutions primarily due to the lack of collision-aware planning for within-aisle goal reordering, then by outside-group agents.}
    \label{fig:worse_solutions}
    \vspace{-1em}
\end{figure}

\smallskip
\noindent\textbf{Rejected Repair Mechanisms} We next examine why proposed repairs can result in less efficient solutions that are ultimately rejected by LSR. We visualized $300$ randomly selected worse proposals ($n=150$ per repair method) and classified them into three causes: (1) interactions with agents outside the repaired group, (2) poor within-aisle goal reordering, or (3) a combination of both factors. Figure~\ref{fig:worse_solutions} shows the distribution of these causes.

For $\text{LSR}_{\text{MCR}}$, all worse proposals arise from interactions with outside-group agents. Although $\text{LSR}_{\text{MCR}}$ finds the minimum-cost assignment for the repaired group, it does not consider agents outside that group during reassignment. A locally beneficial goal change can therefore introduce congestion when the full system is replanned.
For $\text{LSR}_{\text{ECR}}$, worse proposals arise primarily from poor within-aisle goal reordering ($60.7\%$), reflecting its lack of collision-aware planning. Interactions with outside-group agents account for another $31.3\%$, while the remaining $8.0\%$ result from a combination of both causes.

\smallskip
\noindent\textbf{Summary:} $\text{LSR}_{\text{MCR}}$ produces useful goal reassignments substantially more often than $\text{LSR}_{\text{ECR}}$, largely because its collision-aware cost allows it to resolve conflicts both within and between aisles. Its remaining failures arise from interactions beyond the locally repaired group. In contrast, $\text{LSR}_{\text{ECR}}$ is limited both by these outside-group interactions and by its collision-unaware goal reassignment.

\subsection{Repair Computation Time}

\begin{figure}[h]
    \centering
    \includegraphics[width=0.8\linewidth]{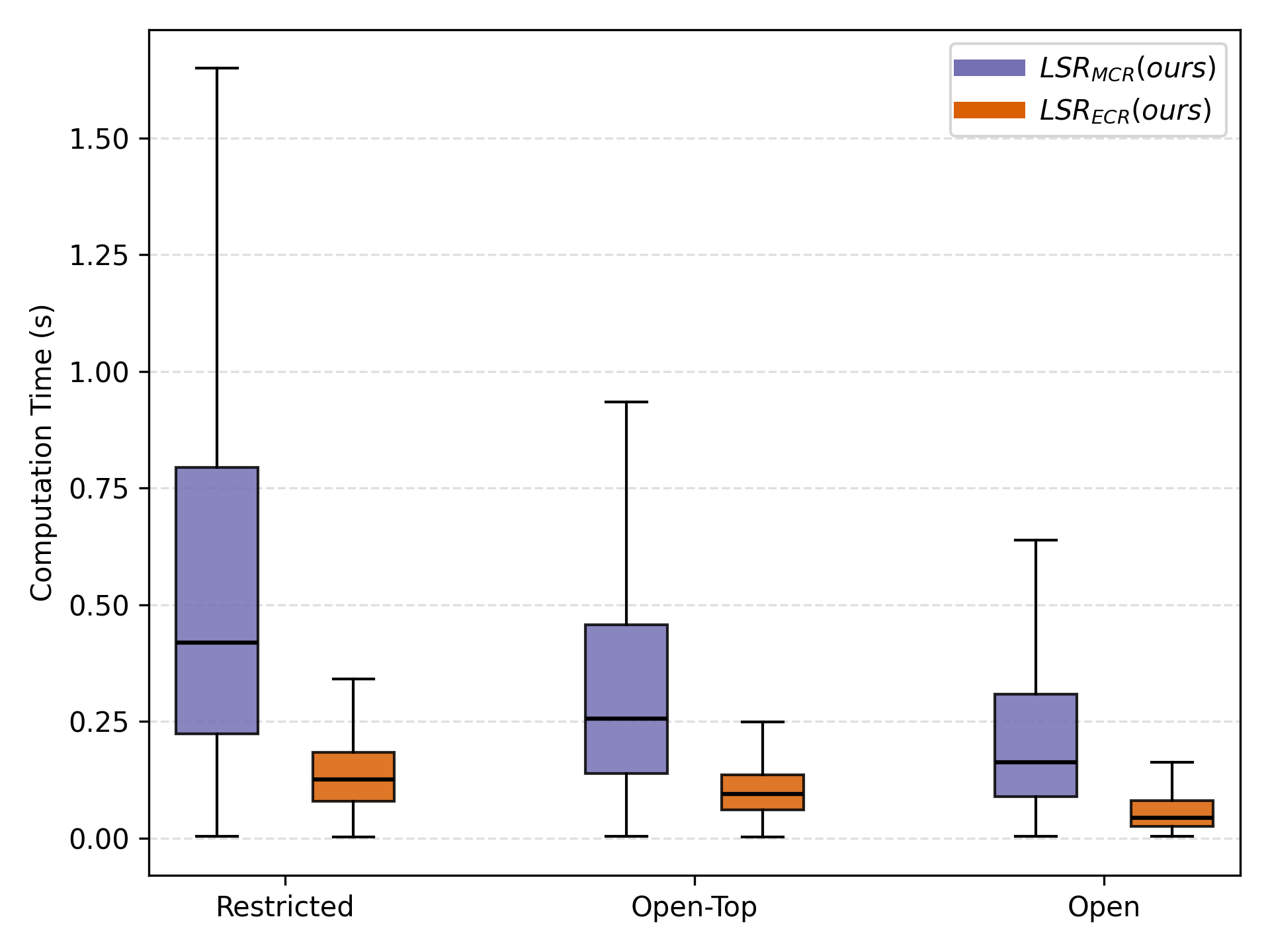}
    \caption{Repair Computation Times (seconds) for $\text{LSR}_{\text{ECR}}$ (orange) and $\text{LSR}_{\text{MCR}}$ (purple) across Restricted (left), Open-Top (middle), and Open maps (right). We note $\text{LSR}_{\text{ECR}}$ substantially reduced computation time compared to $\text{LSR}_{\text{MCR}}$.}
    \label{fig:computation_time}
    \vspace{-1em}
\end{figure}

Figure $\ref{fig:computation_time}$ reports the computation time required by our repair methods for each repair cycle (i.e., time to repair $A_{group}$). The median computation time for $\text{LSR}_{\text{MCR}}$ is $0.42$s, $0.26$s, and $0.16$s for the restricted, open-top, and open environments, respectively. In contrast, $\text{LSR}_{\text{ECR}}$ has reported median computation times of $0.13$s, $0.09$s, and $0.04$s for the three map modes, respectively. 

\textbf{Summary:} There is a trade-off between repair methods: $\text{LSR}_{\text{MCR}}$ achieves higher throughput at the cost of increased computation time, whereas $\text{LSR}_{\text{ECR}}$ attains slightly lower throughput with substantially reduced computational cost.

\section{Conclusion} 
\label{sec:conclusion}
This paper formalizes the Suboptimality Detection and Repair (SDR) problem, in which common suboptimal behaviors produced by an existing MAPD solution are identified and repaired to improve system performance. To solve the SDR problem, we introduce Local Solution Repair (LSR), which identifies subsets of agents that exhibit and contribute to suboptimal behavior, then reassigns agents' goals to nearby equivalent locations. We introduce two LSR variants to compute new goal assignments: Estimated-Cost Repair ($\text{LSR}_{\text{ECR}}$) uses estimated path costs to candidate goal locations and assigns them using the Hungarian Algorithm \cite{kuhn1955hungarian}, and MAPF-Cost Repair ($\text{LSR}_{\text{MCR}}$) uses a Branch-and-Bound algorithm to reassign the group of agents to new locations by computing collision-free path plans during the tree search. Experimental results show that $\text{LSR}_{\text{ECR}}$ and $\text{LSR}_{\text{MCR}}$ improve throughput by upwards of $3.7\%$ and $5.9\%$ respectively by reducing suboptimal actions by $4.4\%$ and $14.6\%$.

More broadly, this work demonstrates that scalability and performance in MAPD need not be achieved solely through tighter coupling of task allocation and motion planning. Instead, lightweight, runtime repair mechanisms such as LSR offer a complementary and practical alternative that preserves the efficiency of existing solvers while correcting their most impactful inefficiencies. Future work will explore adaptive detection thresholds, additional classes of suboptimal behavior, and extensions to higher robot densities, more complex task structures, and real-world deployments.

\bibliographystyle{plainnat}
\bibliography{references}

\begin{thebibliography}{17}
\providecommand{\natexlab}[1]{#1}
\providecommand{\url}[1]{\texttt{#1}}
\expandafter\ifx\csname urlstyle\endcsname\relax
  \providecommand{\doi}[1]{doi: #1}\else
  \providecommand{\doi}{doi: \begingroup \urlstyle{rm}\Url}\fi

\bibitem[Barer et~al.(2014)Barer, Sharon, Stern, and Felner]{barer2014suboptimal}
M.~Barer, G.~Sharon, R.~Stern, and A.~Felner.
\newblock \href{https://ojs.aaai.org/index.php/SOCS/article/view/18315}{Suboptimal variants of the conflict-based search algorithm for the multi-agent pathfinding problem}.
\newblock In \emph{SoCS}, volume~5, 2014.

\bibitem[Chen et~al.(2021)Chen, Alonso-Mora, Bai, Harabor, and Stuckey]{chen2021integrated}
Z.~Chen, J.~Alonso-Mora, X.~Bai, D.~D. Harabor, and P.~J. Stuckey.
\newblock \href{https://ieeexplore.ieee.org/stamp/stamp.jsp?arnumber=9410352}{Integrated task assignment and path planning for capacitated multi-agent pickup and delivery}.
\newblock \emph{IEEE RA-L}, 6\penalty0 (3), 2021.

\bibitem[Gerkey and Matari{\'c}(2004)]{gerkeysurvey}
B.~P. Gerkey and M.~J. Matari{\'c}.
\newblock \href{https://journals.sagepub.com/doi/abs/10.1177/0278364904045564}{A formal analysis and taxonomy of task allocation in multi-robot systems}.
\newblock \emph{IJRR}, 23\penalty0 (9), 2004.

\bibitem[Grenouilleau et~al.(2019)Grenouilleau, Van~Hoeve, and Hooker]{grenouilleau2019multi}
F.~Grenouilleau, W.~J. Van~Hoeve, and J.~N. Hooker.
\newblock \href{https://ojs.aaai.org/index.php/ICAPS/article/view/3474}{A multi-label A* algorithm for multi-agent pathfinding}.
\newblock In \emph{ICAPS}, volume~29, 2019.

\bibitem[Jiang et~al.(2025)Jiang, Wang, Veerapaneni, Duhan, Sartoretti, and Li]{jiang2025deploying}
H.~Jiang, Y.~Wang, R.~Veerapaneni, T.~Duhan, G.~Sartoretti, and J.~Li.
\newblock \href{https://ieeexplore.ieee.org/abstract/document/11127445}{Deploying ten thousand robots: Scalable imitation learning for lifelong multi-agent path finding}.
\newblock In \emph{ICRA}. IEEE, 2025.

\bibitem[Korsah et~al.(2013)Korsah, Stentz, and Dias]{korsahsurvey}
G.~A. Korsah, A.~Stentz, and M.~B. Dias.
\newblock \href{https://journals.sagepub.com/doi/abs/10.1177/0278364913496484}{A comprehensive taxonomy for multi-robot task allocation}.
\newblock \emph{IJRR}, 32\penalty0 (12), 2013.

\bibitem[Kuhn(1955)]{kuhn1955hungarian}
H.~W. Kuhn.
\newblock \href{https://link.springer.com/chapter/10.1007/978-3-540-68279-0_2}{The Hungarian method for the assignment problem}.
\newblock \emph{Naval research logistics quarterly}, 2\penalty0 (1-2), 1955.

\bibitem[Li et~al.(2021)Li, Tinka, Kiesel, Durham, Kumar, and Koenig]{li2021lifelong}
J.~Li, A.~Tinka, S.~Kiesel, J.~W. Durham, TK~Satish Kumar, and S.~Koenig.
\newblock \href{https://ojs.aaai.org/index.php/AAAI/article/view/17344}{Lifelong multi-agent path finding in large-scale warehouses}.
\newblock In \emph{AAAI}, volume~35, 2021.

\bibitem[Liu et~al.(2019)Liu, Ma, Li, and Koenig]{liu2019task}
M.~Liu, H.~Ma, J.~Li, and S.~Koenig.
\newblock \href{https://dl.acm.org/doi/abs/10.5555/3306127.3331816}{Task and path planning for multi-agent pickup and delivery}.
\newblock In \emph{AAMAS}, 2019.

\bibitem[Ma et~al.(2017)Ma, Li, Kumar, and Koenig]{ma2017lifelong}
H.~Ma, J.~Li, T.K.~Satish Kumar, and S.~Koenig.
\newblock \href{https://dl.acm.org/doi/abs/10.5555/3091125.3091243}{Lifelong Multi-Agent Path Finding for Online Pickup and Delivery Tasks}.
\newblock AAMAS, 2017.

\bibitem[Ma et~al.(2019)Ma, Harabor, Stuckey, Li, and Koenig]{ma2019searching}
H.~Ma, D.~Harabor, P.~J. Stuckey, J.~Li, and S.~Koenig.
\newblock \href{https://ojs.aaai.org/index.php/AAAI/article/view/4758}{Searching with consistent prioritization for multi-agent path finding}.
\newblock In \emph{AAAI}, volume~33, 2019.

\bibitem[Okumura(2023)]{okumura2023lacam}
K.~Okumura.
\newblock Lacam: Search-based algorithm for quick multi-agent pathfinding.
\newblock In \emph{AAAI}, volume~37, 2023.

\bibitem[Pisinger and Ropke(2018)]{pisinger2018large}
D.~Pisinger and S.~Ropke.
\newblock \href{https://link.springer.com/chapter/10.1007/978-3-319-91086-4_4}{Large neighborhood search}.
\newblock In \emph{Handbook of metaheuristics}. Springer, 2018.

\bibitem[Schneider et~al.(2026)Schneider, Chen, Gu, Lian, Hutchinson, and Chernova]{schneider2026}
E.~Schneider, J.~Chen, T.~Gu, K.~Lian, S.~Hutchinson, and S.~Chernova.
\newblock Many-to-many multi-agent pickup and delivery, 2026.
\newblock URL \url{https://arxiv.org/abs/2605.07835}.

\bibitem[Sharon et~al.(2015)Sharon, Stern, Felner, and Sturtevant]{sharon2015conflict}
G.~Sharon, R.~Stern, A.~Felner, and N.~R. Sturtevant.
\newblock \href{https://ojs.aaai.org/index.php/AAAI/article/view/8140}{Conflict-based search for optimal multi-agent pathfinding}.
\newblock \emph{Artificial intelligence}, 219, 2015.

\bibitem[Silver(2005)]{silver2005cooperative}
D.~Silver.
\newblock \href{https://ojs.aaai.org/index.php/AIIDE/article/view/18726}{Cooperative pathfinding}.
\newblock In \emph{AAAI}, volume~1, 2005.

\bibitem[Xu et~al.(2022)Xu, Li, Koenig, and Ma]{LNS-PBS}
Q.~Xu, J.~Li, S.~Koenig, and H.~Ma.
\newblock \href{https://ieeexplore.ieee.org/abstract/document/9981785}{Multi-goal multi-agent pickup and delivery}.
\newblock In \emph{IROS}. IEEE, 2022.

\end{thebibliography}

\end{document}